%% file: main.tex
\PassOptionsToPackage{dvipsnames}{xcolor}
\PassOptionsToPackage{hypertexnames=false}{hyperref}
\documentclass[10pt]{article}
\usepackage[final]{scai}
\usepackage[authoryear,sort&compress,round]{natbib}

\graphicspath{}

\input{math_commands.tex}

\input{preamble.tex}

\title{FitText: Evolving Agent Tool Ecologies via Memetic Retrieval}

\contribnote{equal}{Equal contribution.}
\contribnote{corr}{\ddag}{Corresponding author.}

\scaiauthor{1,equal,corr}{Kyle Zheng}
\scaiauthor{1,equal}{Han Zhang}
\scaiauthor{1}{Renliang Sun}
\scaiauthor{1}{Chenchen Ye}
\scaiauthor{1}{Wei Wang}

\affiliation{1}{University of California, Los Angeles}
\scaikeywords{Tool Retrieval, LLM Agents, Test-Time Scaling, Memetic Algorithms, Agentic Retrieval}
\correspondence{Kyle Zheng}{kylezheng@g.ucla.edu}

\begin{abstract}
\input{abstract}
\end{abstract}

\begin{document}
\maketitle

\section{Introduction}
\input{introduction}

\section{Problem Statement}
\label{sec:problem}
\input{problem}

\section{Related Work}
\input{related_work}

\section{Method}
\input{method}

\section{Experiments}
\label{sec:experiments}
\input{experiments}
\input{mainline_limits}

\section{Conclusion}
\input{conclusion}

\bibliographystyle{plainnat}

\input{main.bbl}
\input{ai_disclosure}

\makeappendixtitle
\input{appendix}

\end{document}

%% file: math_commands.tex
\usepackage{amsmath,amsfonts,bm}

\def\eqref#1{equation~\ref{#1}}

\def\1{\bm{1}}

\DeclareMathAlphabet{\mathsfit}{\encodingdefault}{\sfdefault}{m}{sl}
\SetMathAlphabet{\mathsfit}{bold}{\encodingdefault}{\sfdefault}{bx}{n}

\newcommand{\Ret}{\textsc{Ret}}
\newcommand{\LLM}{\textsc{LLM}}
\newcommand{\Pop}{\mathcal{P}}
\newcommand{\Mem}{\mathcal{M}}
\newcommand{\Corpus}{\mathcal{T}}

%% file: abstract.tex
Efficient reasoning is not only a matter of shortening an answer trace; for tool-using agents, it also depends on whether the agent is reasoning over the right action space. As API ecosystems scale to tens of thousands of endpoints, the semantic gap between user requests and tool documentation makes this problem concrete: static retrieval from the initial query can fail before planning begins, and stronger planning alone cannot recover a missing tool. We study this problem as budgeted test-time retrieval and introduce \ourmethod, a training-free framework that makes the tool interface revisable during execution by generating, refining, and evolving natural-language pseudo-tool descriptions as retrieval probes. \ourmethod\ supports serial refinement, parallel exploration, and Memetic Retrieval, which adds evolutionary selection, local refinement, and tool memory to avoid redundant search. On StableToolBench (16{,}464 APIs), Memetic \ourmethod reaches an $84.3\%$ pooled pass rate, improving $+26.7$ points over static retrieval, $+22.2$ over Single-Pass, $+23.2$ over Re-Invoke, and $+27.5$ over Xu-style root refinement. It leads on every evaluated current model, with gains growing alongside model capability, and produces the largest improvements on ambiguous multi-tool tasks where dynamic re-retrieval restores correct candidates after early mistakes. At 40-way concurrency, parallel population execution keeps batched wall-clock at $1.01\times$ Single-Pass despite the added search work.

%% file: introduction.tex
Large language models (LLMs) demonstrate strong reasoning and planning capabilities \citep{minaee2024large}, yet remain limited in their ability to interact with the external world through tools and APIs.
To overcome this limitation, tool-augmented LLM agents have been introduced, enabling models to invoke APIs, databases, and external environments for more complex and grounded task execution \citep{qin2024tool, qu2025tool}.

A fundamental challenge in such systems is identifying which tools to use for a given task, known as \textbf{tool retrieval}.
Existing approaches fall into two paradigms, both with critical limitations:
\textbf{(i)~Schema injection}, which provides agents with a manually curated, fixed set of candidate tools in the prompt context~\citep{huang2023metatool, shen2023hugginggpt, paranjape2023art}, as standard benchmarks assume~\citep{guo2024stabletoolbench, qintoolllm}; this inflates the context with irrelevant schemas, and manual curation becomes unrealistic as tool ecosystems grow to tens of thousands of APIs~\citep{qintoolllm}.
\textbf{(ii)~Static root-query retrieval}, which selects tools once before execution from the initial user query~\citep{qu2024towards}; but user intent is amorphous and a query alone may not capture information needs~\citep{asai2023taskaware}, making one-shot retrieval brittle.
Both paradigms assume all required tools can be determined upfront, yet agents often discover the need for new tools \emph{during} reasoning as their understanding evolves.
This motivates \textbf{dynamic retrieval}, where tool discovery is integrated into reasoning, letting the agent decide \emph{when} and \emph{what} to retrieve adaptively throughout execution.

Compounding this, prior methods overlook the \emph{semantic gap} between queries and tool descriptions: queries are high-level and goal-oriented (e.g., Figure~\ref{fig:dynamic-retrieval-example}, ``My flight was cancelled, help me get a refund’’), while tool descriptions are functional (e.g., ``Get the current status of a flight’’). This gap widens in large, evolving API catalogs whose relevant functionality cannot be anticipated at query time, and under general-purpose embeddings not trained to match task intent against tool metadata.

To address these challenges, we propose \ourmethod, an LLM-guided dynamic tool retrieval framework (Figure~\ref{fig:dynamic-retrieval-example}). Instead of using the user query directly, the LLM reasons about the task and produces a \emph{pseudo-tool description}: a natural-language hypothesis of the needed tool's functionality that enables retrieval in a semantically aligned space. Viewed this way, a pseudo-tool description is a \emph{revisable hypothesis} about which tool should exist, and retrieval against the corpus is a cheap test of that hypothesis, which the agent revises as evidence arrives.

This parallels query rewriting, hypothetical document generation, and pseudo-relevance feedback in RAG~\citep{li2024dmqrrag, liu2024queryrewrite, gao2022hyde, wang2023query2doc, lavrenko2001relevance}, where a hypothetical answer bridges the query--document gap; but tools are structured with parameters and schemas, must be \emph{executable}, and the agent's evolving context provides a richer refinement signal than a static query. We also adapt corpus-steered query expansion~\citep{lei2024corpussteered} to the tool domain.

We organize the paper around four questions: whether generated tool descriptions improve retrieval quality (RQ1), whether dynamic retrieval improves end-to-end tool use (RQ2), whether the gain comes from recovering candidate availability after an initially wrong tool (RQ3), and how capability plus execution budget govern its quality--efficiency frontier (RQ4).
Our main contributions are summarized as follows:
\begin{itemize}[leftmargin=*,noitemsep,topsep=2pt]
    \item \textbf{RQ1: Generated descriptions reduce the query--tool gap.} Static retrieval treats the user query as a fixed key against a passive index, but intent is amorphous and the right tool is often unknowable at query time. We instead let the LLM search over \emph{revisable hypotheses} of the tool it needs, expressed as pseudo-tool descriptions and scored against the fixed corpus as a cheap fitness signal, with no training data or weight updates.

    \item \textbf{RQ2: Dynamic retrieval improves end-to-end tool use.} We organize the reformulation strategies into a design space spanned by refinement depth and population breadth, whose corners recover Single-Pass, Multi-Turn, and Scattershot. Memetic Retrieval is the full method: a multi-generation evolutionary loop over pseudo-tool descriptions with crossover, mutation, retrieval-fitness selection, and tool memory. On StableToolBench with GPT-5.4-mini, Memetic reaches an $84.3\%$ pooled pass rate: $+26.7$ over static query retrieval, $+23.2$ over Re-Invoke, and $+27.5$ over Xu-style root refinement, with gains concentrated on ambiguous multi-tool tasks.

    \item \textbf{RQ3: The mechanism is candidate recovery, not better first guesses.} On hard G2 tasks, dynamic re-retrieval does not substantially change the first-tool error rate. It instead raises the probability that a correct alternative re-enters the top-5 retrieved set after a wrong first pick, making the downstream solver's candidate frame less brittle.

    \item \textbf{RQ4: Retrieval is the binding constraint and an efficient test-time scaling target.} Across the planning-model variants evaluated, stronger planning does not materially improve pass rate, whereas Memetic's advantage over static retrieval grows $+16.0\!\to\!+18.6\!\to\!+26.7$ points with model capability. Relative to Single-Pass, Memetic adds $22.2$ pass points while retaining $1.01\times$ batched wall-clock through parallel population search.
\end{itemize}

%% file: problem.tex
\begin{figure*}[t]
  \centering
  \includegraphics[width=0.92\textwidth]{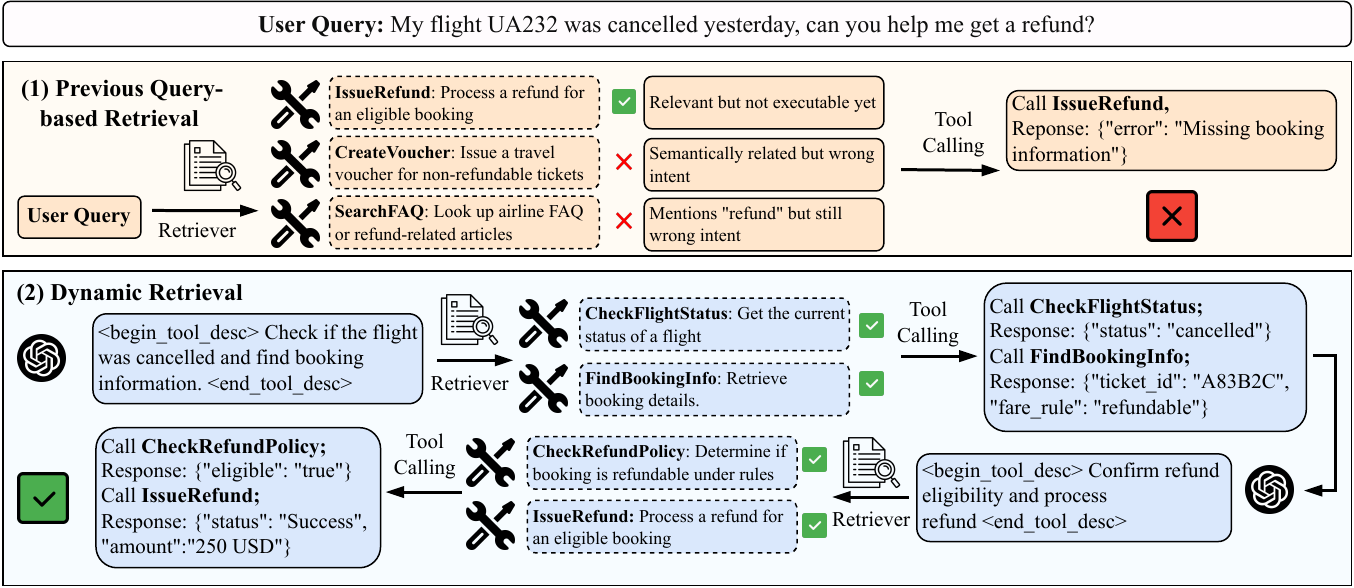}
  \caption{Static query retrieval freezes the action space before execution: it exposes \texttt{IssueRefund}, but omits the booking, flight-status, and policy tools needed to make that refund call executable. Dynamic retrieval revises the schema after the failure, retrieves those prerequisite tools, and then returns to \texttt{IssueRefund} with the required information.}
  \label{fig:dynamic-retrieval-example}
\end{figure*}

We frame tool retrieval as a \textbf{zero-shot retrieval and decision-making problem}: given a task goal derived from user query $q$, the agent must identify, possibly across multiple steps, the most relevant tools from $\Corpus$ without task-specific training data or hardcoded retrieval pipelines.

Formally, the agent generates a sequence of actions $a_1, a_2, \ldots, a_T$ where $a_t \in \mathcal{A}$.
At each step $t$, the available actions depend on the \emph{function schema} $\mathcal{F}_t \subseteq \Corpus$, the subset of tools currently exposed to the agent from the full tool corpus $\Corpus$.
In long-horizon tasks, retrieval errors at early steps compound across the action sequence~\citep{song2026reasoning}, making the quality of $\mathcal{F}_t$ a potential bottleneck for task success, a claim our experiments test directly (\S\ref{sec:experiments}).
The upper bound on this quality is the set of \emph{oracle tools}: the ground-truth gold tool set handed to the agent (as in prior work, directly or in the user prompt), which is unrealistic in deployment since neither the system nor the user knows the gold set a priori.

\paragraph{Strategy trigger.}
\label{sec:trigger}
The framework permits retrieval to be invoked during reasoning: at any step $t$, the agent may emit a pseudo-tool description $d_t$, causing the system to update the function schema $\mathcal{F}_{t+1} \gets \mathcal{F}_t \cup \mathrm{Retrieve}(d_t, \Corpus)$.
The trigger mechanism and its implementation are detailed in \S\ref{sec:method-framework}.

\paragraph{Desiderata.}
We seek a tool retrieval framework satisfying the following properties:
\textbf{(i)~Model-agnostic:} the framework must operate with any base LLM without architecture-specific modifications;
\textbf{(ii)~Training-free:} it must not require fine-tuning the retriever or LLM, enabling deployment with API-only models;
\textbf{(iii)~Orthogonal:} it must compose with advances in agentic planning and stronger base models as a modular, plug-and-play capability.

%% file: related_work.tex
\subsection{Tool-Augmented Large Language Models}
Tool-augmented LLM agents enhance reasoning through external tool invocation. Early systems connected models to browsers and APIs~\citep{nakano2021webgpt, song2023restgpt, wang2024executable}, and ReAct~\citep{yao2022react} established observation-action reasoning loops.
Subsequent work splits into training-based methods that fine-tune on tool-call datasets~\citep{schick2023toolformer, patil2024gorilla} and training-free approaches that inject tool schemas into prompts for in-context selection~\citep{huang2023metatool}.
Both, however, assume oracle tools are provided, which is unrealistic.

\subsection{Tool Retrieval for LLM Agents}
Tool retrieval identifies the top-$K$ most relevant tools from a large repository for a given query, the foundation for effective tool use in LLM agents.
While prior research concentrates on tool \emph{usage} and calling (e.g., planning-centric action sequencing~\citep{zhuang2023toolchain}), the upstream retrieval bottleneck remains underexplored, with selection errors propagating through inference trees~\citep{chen2025advancing}; on-demand retrieval has been shown to scale better on long-horizon tasks even without training~\citep{li2025deepagent}.
Early approaches adopt conventional information retrieval: term-based models (TF--IDF~\citep{sparck1972statistical}, BM25~\citep{robertson2009probabilistic}) rely on sparse lexical overlap and fail when users describe functionality in natural, task-specific language.

To capture deeper semantics, recent work shifts toward dense retrieval inspired by RAG \citep{lewis2020retrieval}, encoding queries and tools into a shared embedding space. Tool2Vec \citep{moon2024efficient} models user invocation patterns; AnyTool \citep{du2024anytool} and ToolRerank \citep{zheng2024toolrerank} perform coarse-to-fine selection across category-tool-API levels; COLT \citep{qu2024colt} adds completeness-oriented reranking; Re-Invoke \citep{chen2024re} rewrites invocations; and MCP-Zero \citep{fei2025mcp} lets LLMs actively request tools, but remains conceptual with minimal validation.
ToolDreamer \citep{sengupta2025tooldreamer} instills LLM reasoning into retrievers via training, and \citet{xu2024enhancing} use iterative LLM feedback---goals we reach at inference time without parameter updates. ToolGen \citep{wang2025toolgen} unifies retrieval and calling generatively, whereas we decouple tool discovery from invocation.
Despite these advances, existing frameworks share fundamental limitations.
\textbf{(1) Static retrieval:} most retrieve tools only once, preventing adaptation as reasoning evolves (\S1).
\textbf{(2) Structure-dependent designs:} many assume hierarchical taxonomies or pre-collected invocation patterns, restricting them in unstructured environments without those resources. Our experiments do use normalized tool descriptions: full StableToolBench documentation is summarized once before evaluation, identically for every retrieval method (\S\ref{sec:setup}; Appendix~\ref{app:setup}).
\textbf{(3) Semantic gap:} direct query-to-tool matching yields poor recall (\S1); prior work aligns tool embeddings toward queries, while we invert this direction via pseudo-tool generation.

\paragraph{Relation to recent tool-retrieval methods.}
Unlike \ourmethod\ (training-free, cold-start, and re-retrieving mid-execution), each recent method rests on an assumption we avoid: a corpus of past solved trajectories (SEER~\citep{cui2025seer}); a single pre-test-time tool prediction akin to static retrieval (Less-is-More~\citep{paramanayakam2024lessismore}); or a trained discriminator/encoder over a global tool or dependency graph (Tool Graph Retriever~\citep{gao2025tgr}, ControlLLM~\citep{liu2023controlllm}, COLT~\citep{qu2024colt}). None evolves query-side probes dynamically within a single episode.

\paragraph{Connection to prompt optimization.}
Automatic prompt optimization searches discrete textual spaces~\citep{li2025promptsurvey}, recently via LLM-driven evolutionary operators and exploration--exploitation trade-offs~\citep{agrawal2025gepa, cui2025see}; these optimize instruction prompts against a downstream objective, whereas we search pseudo-tool descriptions whose fitness is retrieval quality against the tool corpus.

\paragraph{Test-time policy improvement.}
\label{par:test-time-policy}
A growing body of work spends inference compute to improve an LLM policy without weight updates: marginalizing reasoning paths~\citep{wang2023selfconsistency}, iterative self-refinement~\citep{madaan2023selfrefine}, verbal-feedback memory~\citep{shinn2023reflexion}, search over reasoning traces~\citep{yao2023treeofthoughts}, and compute-optimal test-time allocation~\citep{snell2024scaling}. Almost all target the \emph{answer} policy: how the model generates or selects the final response. \ourmethod\ applies the same principle one level earlier, to the \emph{retrieval} sub-policy: it evolves revisable pseudo-tool descriptions against the corpus as a fitness signal, improving the tool interface the downstream policy depends on. Because the improvement grows with base-model capability, \ourmethod\ benefits from test-time scaling~\citep{snell2024scaling} rather than inheriting a frozen retriever's ceiling.

%% file: method.tex
\begin{figure}[t]
    \centering
    \includegraphics[width=0.92\textwidth]{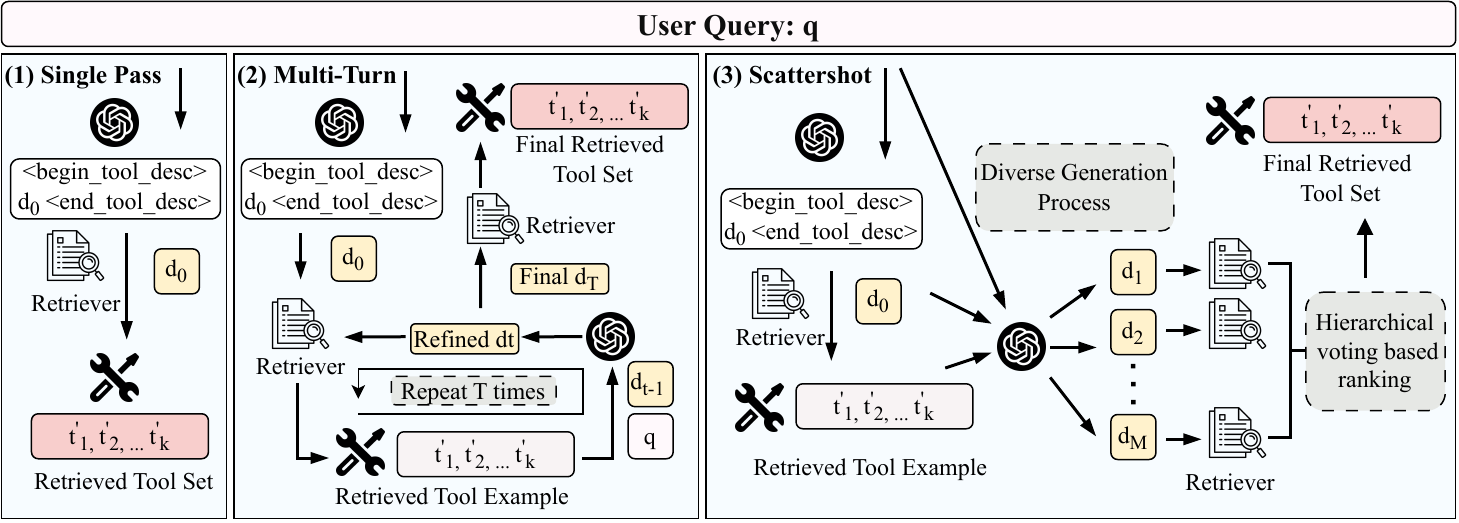}
    \caption{The boundary cases of our design space. Each strategy generates \emph{pseudo-tool descriptions} as retrieval probes, differing in search depth and population breadth: \textit{Single-Pass} (one probe), \textit{Multi-Turn} (deepen one lineage), and \textit{Scattershot} (broaden a population, then vote); \emph{Memetic} (Fig.~\ref{fig:memetic-workflow}) adds evolutionary selection at the apex.}
    \label{fig:retrieval_framework}
\end{figure}

\subsection{Overview}
\label{sec:method-overview}

We frame tool retrieval as \textbf{tool exploration}: the API catalog is a large, heterogeneous \emph{tool ecology} whose niches are defined by user queries, and the agent searches it during reasoning by generating \emph{pseudo-tool descriptions}---intermediate hypotheses about needed functionality that serve as retrieval probes (\S\ref{sec:method-framework}).

We organize these strategies as points in \textbf{one design space} over pseudo-tool descriptions, parameterized by population size $N$, generations $G$, and which operators are enabled: retrieval-fitness selection, meaning-level crossover, mutation, local-search refinement, and a tool-memory penalty. \emph{Memetic Retrieval} is the full instantiation with all operators on (\S\ref{sec:method-memetic}); the simpler strategies are \emph{boundary cases} that each disable some operators: \emph{Multi-Turn Refinement} ($N{=}1$, single-lineage refinement, no crossover or memory penalty; \S\ref{sec:method-multiturn}), \emph{Scattershot} ($G{=}1$, diverse probes voted, no selection or local search; \S\ref{sec:method-scattershot}), and \emph{Single-Pass} (the joint limit $N{=}1, G{=}1$, all evolutionary operators off). These are distinct corners of the space, not partial runs of one algorithm;\footnote{The method name denotes the design space and its fully realized instantiation, not every strategy.} they span fast (Single-Pass) to thorough (Memetic)~\citep{guo2025evoprompt}, with an operator-by-operator comparison in Appendix Table~\ref{tab:design_space}.

\paragraph{Search budget.}
We use \emph{budget} to mean a fixed configuration $\mathbf{b}=(T,S,N,G)$ chosen before evaluation: serial refinement depth, parallel fan-out, population size, and number of generations. The framework does not learn a per-query router that selects among these budgets. We distinguish \emph{total work} (LLM generations and retrieval calls) from \emph{serialized depth}; parallel branches can shorten elapsed time without reducing tokens or calls. The experiments therefore compare discrete compute--quality points rather than claiming adaptive budget allocation.

\paragraph{Tool Retrieval as Inference-Time Hypothesis Evolution.}
The strategies above share a single view: each \emph{pseudo-tool description} is a mutable hypothesis that ``a tool with function $d$ exists in $\Corpus$.'' Retrieval supplies an unsupervised, corpus-grounded confidence proxy for that hypothesis, so the agent searches over hypothesized tool functionality (the document side of the semantic gap) rather than over rewrites of the query's surface form. A hypothesis is better supported when its retrieval mass concentrates on concrete tools; it is redundant when its embedding falls in an already explored region of description space. This is generate-then-retrieve in the spirit of hypothetical document generation~\citep{gao2022hyde}, made \emph{iterative} (Eq.~\ref{eq:mt-refine}), \emph{evolutionary} (Alg.~\ref{alg:memetic}), and \emph{memory-penalized} (Eq.~\ref{eq:fitness}). Retrieval confidence is not a ground-truth oracle for semantic correctness.

Two properties make this an inference-time mechanism. First, hypotheses are \emph{revisable}: retrieval exposes the vocabulary and neighborhood of tools that actually exist, so a poorly formed $d_0$ need not be fatal. After execution feedback, dynamic re-retrieval can restore missing tool availability and create a self-correction path that frozen retrieval cannot express (\S\ref{sec:recovery}).

Second, the KL posterior-drift penalty (Eq.~\ref{eq:fitness}) discourages candidates that crowd already explored regions of description space. \ourmethod\ is thus a test-time policy-improvement method aimed at the \emph{retrieval interface} rather than the answer policy, with no weight update (\S\ref{par:test-time-policy}).

\subsection{Dynamic Tool Retrieval Framework}
\label{sec:method-framework}

When a strategy trigger fires (\S\ref{sec:trigger}), the system extracts one or more \textbf{pseudo-tool descriptions} $d = \LLM(q, \mathcal{H})$ from the agent's output $a_t$, where $\mathcal{H}$ is the trajectory history maintained by the depth-first search decision tree (DFSDT) reasoning loop~\citep{qintoolllm}, whose backtracking rewrites earlier steps. Concretely, retrieval fires whenever the agent's emitted thought contains a delimited pseudo-tool block that is not a near-duplicate of a previously explored intent (memory-based deduplication; integration details and prompt templates in Appendices~\ref{app:dfsdt} and~\ref{app:prompts}). Because generating a tool hypothesis requires reasoning about expected inputs and outputs, each pseudo-tool implicitly encodes a finer task plan than cluster-based tool selection~\citep{liu2025toolplanner}.

\paragraph{Retrieval mechanism.}
Given a pseudo-tool description $d$ and the tool corpus $\Corpus = \{t_1, \ldots, t_n\}$, we score each tool by embedding cosine similarity $\mathrm{score}(d, t_i) = \cos(\mathbf{e}(d), \mathbf{e}(t_i))$, where $\mathbf{e}(\cdot)$ is a sentence embedding function (SimCSE~\citep{gao2021simcse}, RoBERTa-large variant), and retrieve the top-$k$ tools $R = \Ret(d, \Corpus, k)$ by descending score. Retrieved tools are injected into the function schema $\mathcal{F}_{t+1} \gets \mathcal{F}_t \cup R$ for subsequent steps.

\paragraph{Single-Pass Retrieval.}
\label{sec:method-singlepass}
The initial description $d_0 = \LLM(q, \mathcal{H})$ is used directly for one retrieval step with no refinement. Even a single description bridges the semantic gap---it operates in a space closer to tool documentation than the user query---but cannot correct misalignment between $d_0$ and the corpus after that first retrieval.

\subsection{Multi-Turn Refinement}
\label{sec:method-multiturn}

When $d_0$ misaligns with actual tool representations in $\Corpus$, each retrieval step returns concrete tools that \emph{actually exist}, exposing the vocabulary, structure, and granularity of real tool descriptions. Drawing on iterative self-refinement~\citep{madaan2023selfrefine}, we use this grounding signal to progressively refine the hypothesis. With $d^{(0)} = d_0$, at each refinement turn $t = 1, \ldots, T$:
\begin{align}
R^{(t)} &= \Ret\!\big(d^{(t-1)},\, \Corpus,\, k\big), \label{eq:mt-retrieve} \\
\mathcal{B}^{(t)} &= \mathcal{B}^{(t-1)} \cup \big\{\mathrm{blurb}(r) : r \in R^{(t)}\big\}, \label{eq:mt-exemplars} \\
d^{(t)} &= \LLM_{\mathrm{refine}}\!\big(d^{(t-1)},\, \mathcal{B}^{(t)},\, q\big), \label{eq:mt-refine}
\end{align}
where $\mathcal{B}^{(t)}$ is the accumulated (deduplicated) exemplar set and $\mathrm{blurb}(r)$ a normalized summary of tool $r$; the final retrieval uses $d^{(T)}$. The refinement temperature $\tau_{\mathrm{refine}}$ is low (Table~\ref{tab:hyperparams}) for focused, incremental refinement. Still, Multi-Turn follows a single trajectory that can converge to a local optimum when $d_0$ is poorly formulated.

\subsection{Scattershot Retrieval}
\label{sec:method-scattershot}

To overcome the single-trajectory limitation of Multi-Turn, Scattershot Retrieval samples $S$ diverse pseudo-tool descriptions at high temperature, so that at least one lands near each required tool, then aggregates their retrieval results through voting.

From a seed retrieval $R_0 = \Ret(d_0, \Corpus, k)$ with exemplars $\mathcal{E} = \{\mathrm{blurb}(r) : r \in R_0\}$, we generate $S$ children in parallel at the elevated temperature $\tau_{\mathrm{scatter}}$ (Table~\ref{tab:hyperparams}):
\begin{equation}
\label{eq:scatter-gen}
d_i \;\sim\; \LLM_{\mathrm{scatter}}\!\big(d_0,\, \mathcal{E};\; \tau_{\mathrm{scatter}}\big), \quad i = 1,\ldots,S.
\end{equation}
Each child retrieves $R_i = \Ret(d_i, \Corpus, k)$; results are aggregated by vote count, mean rank, and mean similarity (Algorithm~\ref{alg:vote}).

\subsection{Memetic Retrieval}
\label{sec:method-memetic}

\begin{figure}[t]
  \centering
  \includegraphics[width=0.90\textwidth]{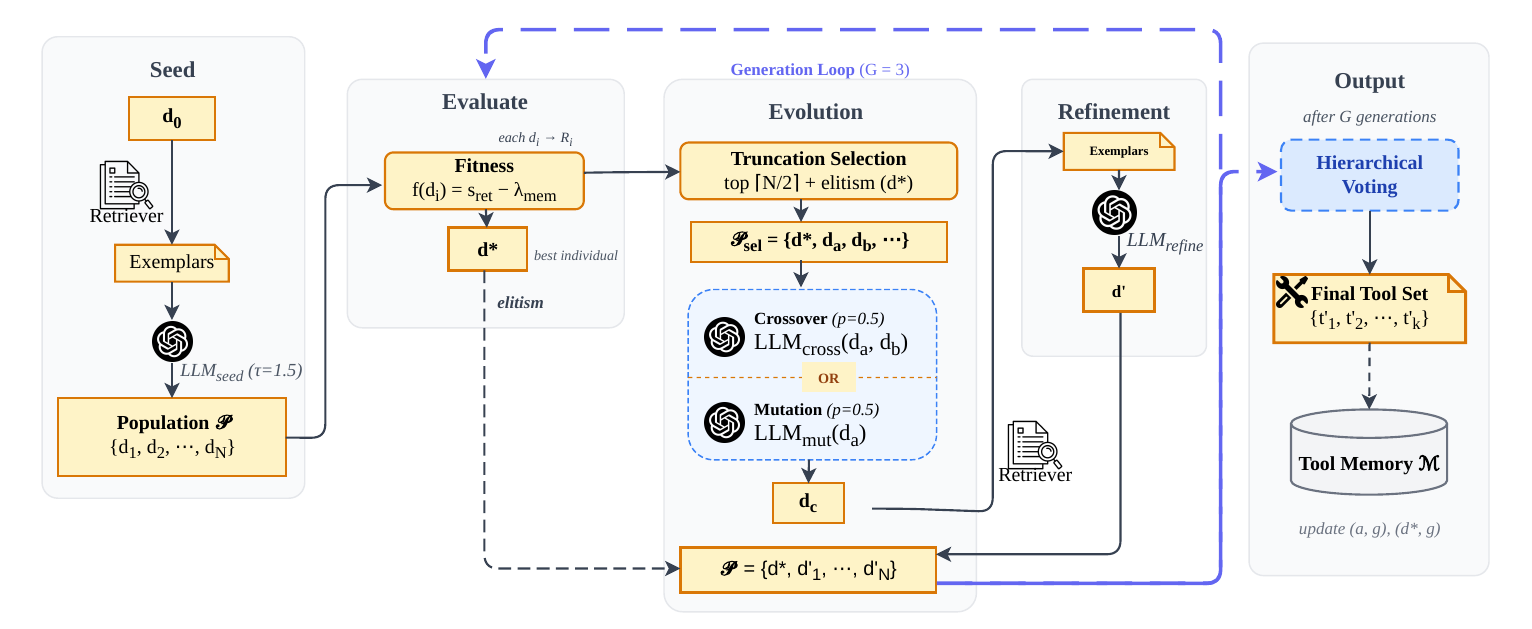}
  \caption{Memetic Retrieval, the apex of the design space. Five phases: \textbf{Seed} an initial population of pseudo-tool descriptions; \textbf{Evaluate} each by retrieval fitness (Eq.~\ref{eq:fitness}); \textbf{Evolution} via truncation selection with meaning-level crossover and mutation; \textbf{Refinement} of offspring through retrieval feedback; \textbf{Output} via hierarchical voting. The loop repeats for $G$ generations, with a tool memory suppressing redundant search.}
  \label{fig:memetic-workflow}
\end{figure}

Scattershot generates a diverse population but discards all information after a single generation. \textbf{Memetic Retrieval} instead treats each pseudo-tool description as a mutable hypothesis---a \textit{meme} in cultural evolution---combining population-based exploration with iterative refinement through a multi-generation evolutionary loop (Alg.~\ref{alg:memetic}; Figure~\ref{fig:memetic-workflow} traces its five phases of Seed, Evaluate, Evolution, Refinement, and Output): competing hypotheses are scored by retrieval fitness, subjected to selection pressure, and recombined through crossover and mutation, with a local-search step refining each offspring against retrieval feedback. The LLM serves as a \emph{semantic} evolutionary operator~\citep{guo2025evoprompt,agrawal2025gepa}, performing meaning-preserving crossover and intent-aware mutation rather than token-level perturbation~\citep{novikov2025alphaevolve}.

\paragraph{Retrieval confidence and fitness.}
Let $s_{(1)}(d_i) \geq \cdots \geq s_{(k)}(d_i)$ be the ordered cosine similarities for $R_i=\Ret(d_i,\Corpus,k)$. The unpenalized retrieval-confidence score used for stopping is
\begin{equation}
\label{eq:retrieval-confidence}
r(d_i)=\alpha s_{(1)}(d_i)+(1-\alpha)\frac{1}{m}\sum_{j=1}^{m}s_{(j)}(d_i),
\end{equation}
with $\alpha=0.7$ and $m=3$. Separately, each pseudo-tool description induces a retrieval distribution over its candidates: a softmax over the similarity scores,
\begin{equation}
\label{eq:retdist}
\pi_{d_i}(t)=\frac{\exp s(t)}{\sum_{t'\in R_i}\exp s(t')},\qquad t\in R_i.
\end{equation}
The fitness ranks descriptions by concentrated retrieval mass while penalizing redundant search through posterior drift:
\begin{equation}
\label{eq:fitness}
f(d_i)=\underbrace{\log\!\!\sum_{t\in\mathrm{top}\text{-}m(R_i)}\!\!\pi_{d_i}(t)}_{\text{observation likelihood}}
-\underbrace{\lambda D_{\mathrm{KL}}\!\left(\mathcal{P}_{\Mem\oplus d_i}\,\middle\|\,\mathcal{P}_{\Mem}\right)}_{\text{memory penalty}}.
\end{equation}
Here $\mathcal{P}_{\Mem}$ is a Gaussian-kernel density over normalized embeddings of previously evaluated descriptions, and $\mathcal{P}_{\Mem\oplus d_i}$ is the density after admitting $d_i$ (Appendix~\ref{app:memory}). The likelihood term rewards descriptions whose strongest retrieved tools receive high mass; the KL term discourages repeatedly exploring an already-covered region. Selection ranks candidates by $f$, while early stopping remains tied to the distinct, unpenalized score: the loop stops when $r(d^*)\geq\theta=0.95$ or reaches $G$. This separation keeps the memory penalty from being interpreted as calibrated confidence.

The fitness form is fixed across the reported experiments; only the search-budget knobs ($N$, $G$, refinement turns $T$, and Scattershot samples $S$) are selected on development data. A matched comparison of token overlap, KL posterior drift, Hellinger, and determinantal penalties selected KL drift for the reported configuration.

\begin{algorithm}[t]
\caption{Memetic Retrieval}\label{alg:memetic}
\footnotesize
\begin{algorithmic}[1]
\Require $q$, $\mathcal{D}_0$, $\Corpus$, $k$, $\Mem$, $N$, $G$, $\theta$, $\tau_{\mathrm{memetic}}$, $p_{\mathrm{cross}}$, $B$ \hfill\emph{values in Table~\ref{tab:hyperparams}}
\Ensure Ranked tool list; updated $\Mem$
\State $\mathit{keys}\gets[\,]$
\For{each ancestor $a \in \mathcal{D}_0$}
  \State $\Pop \gets \textsc{SeedPop}(a, \Corpus, k, N, \tau_{\mathrm{memetic}})$ \hfill\emph{Alg.~\ref{alg:seedpop}}
  \For{$g \gets 1$ to $G$}
    \For{each $d_i \in \Pop$} \;$R_i \gets \Ret(d_i, \Corpus, k)$;\; $f_i \gets f(d_i)$ \hfill\emph{Eq.~\ref{eq:fitness}}
    \EndFor
    \State $d^* \gets \arg\max_i f_i$;\; \textbf{if} $r(d^*) \geq \theta$ or $g = G$ \textbf{then break}
    \State $\Pop_{\mathrm{sel}} \gets \mathrm{top}_{\lceil N/2 \rceil}(\Pop, f)$;\; $\Pop' \gets \{d^*\}$ \hfill\emph{elitism}
    \While{$|\Pop'| < N$} \hfill\emph{crossover or mutation}
      \If{$\mathrm{Uniform}(0,1) < p_{\mathrm{cross}}$} \;$d_a, d_b \gets \mathrm{sample}(\Pop_{\mathrm{sel}}, 2)$;\; $d_c \gets \LLM_{\mathrm{cross}}(d_a, d_b, a;\; \tau_{\mathrm{memetic}})$
      \Else \;$d_a \gets \mathrm{sample}(\Pop_{\mathrm{sel}}, 1)$;\; $d_c \gets \LLM_{\mathrm{mut}}(d_a, a;\; \tau_{\mathrm{memetic}})$
      \EndIf
      \State $\Pop' \gets \Pop' \cup \{d_c\}$
    \EndWhile
    \For{each $d \in \Pop' \setminus \{d^*\}$} \;$d \gets \LLM_{\mathrm{refine}}(a, d, \Ret, q)$ \hfill\emph{local search}
    \EndFor
    \State $\Pop \gets \Pop'$
  \EndFor
  \State $\Mem \mathrel{+}= \{(a, g),\, (d^*, g)\}$;\; $\mathit{keys}.\mathrm{extend}\!\big(\textsc{Vote}(\Pop, \Corpus, k, B)\big)$ \hfill\emph{Alg.~\ref{alg:vote}}
\EndFor
\State \Return $\mathit{keys}$
\end{algorithmic}
\end{algorithm}

%% file: experiments.tex
\subsection{Experimental Setup}
\label{sec:setup}

We evaluate primarily on \textbf{StableToolBench} \citep{guo2024stabletoolbench} (16,464 APIs), a stabilized variant of ToolBench \citep{qintoolllm} for end-to-end agentic tool use, where retrieval quality can limit performance when gold tools are not provided in advance~\citep{shi2025retrieval}. We use \textbf{ToolRet} \citep{shi2025retrieval} (43k tools) as a retrieval-only diagnostic at scale.
We compare \textbf{Query Retrieval} (the static baseline, written \emph{Query (static)} hereafter), \textbf{Single-Pass} (\S\ref{sec:method-singlepass}), \textbf{Multi-Turn Refinement} (\S\ref{sec:method-multiturn}), \textbf{Scattershot Retrieval} (\S\ref{sec:method-scattershot}), and \textbf{Memetic} (\S\ref{sec:method-memetic}) using \emph{GPT-5.4-mini} (headline), \emph{Qwen3.6-35B}, and \emph{Gemma4-31B} as base models.
Query (static) retrieves from the raw user query, whereas Single-Pass retrieves once from one generated pseudo-tool description. The evaluated search menu uses $T\in\{3,5\}$ refinement turns, $S\in\{5,10\}$ parallel samples, and Memetic $N{=}5$, $G{=}3$, with top-$k$ and output budget $B$ both fixed at $5$ (cell-level settings in Appendix~\ref{app:hyperparams}).
On ToolRet, the named model generates and refines pseudo-tool descriptions while the SimCSE retriever is fixed. On StableToolBench, the named suite model is the downstream solver and supplies reformulation calls within that configuration; GPT-5.4-mini judges task success (Appendix~\ref{app:setup}).
StableToolBench uses the same fixed 765 task IDs for every strategy (G1-category 153, G1-instruction 163, G1-tool 158, G2-category 124, G2-instruction 106, G3-instruction 61). This is the benchmark manifest's \emph{solvable} subset label, not a filter selected using our outcomes; each task is judged eight times and pooled rates weight tasks rather than split means.
For each benchmark, the full tool documentation is summarized once by GPT-4.1-mini before retrieval, and the resulting benchmark-specific corpus is shared by all methods; this preprocessing is outside per-query inference.
On ToolRet we report \emph{Global NDCG@5} (computed after merging qrels and results), \emph{Precision@5}, \emph{Recall@5}, and \emph{Comprehensiveness@5} (whether all gold tools appear in the top five); on StableToolBench, \emph{pass rate} is the fraction of tasks the judge marks solved.
StableToolBench organizes tasks into G1 (single-tool), G2 (same-category multi-tool), and G3 (cross-category multi-tool) levels (Appendix~\ref{app:benchmark}).
Full setup details (tool corpus construction, base model configurations, and metric definitions) appear in Appendix~\ref{app:setup}.
The experiments follow the four research questions stated in the introduction: retrieval quality first, end-to-end utility second, recovery mechanism third, and limits plus efficiency last.

\begin{table}[tb]
\centering
\small
\caption{End-to-end pass rates (\%) on StableToolBench (765-task solvable subset, GPT-5.4-mini judge, eight evaluations per task), pooled across splits, against the static \emph{Query} (SimCSE) floor. The $\Delta$ row (Memetic\,$-$\,Query) grows monotonically with base-model capability. Per-split breakdowns appear in Table~\ref{tab:toolbench_gpt_split}.}
\label{tab:toolbench_passrate}
\begin{tabular}{@{}lccc@{}}
\toprule
\textbf{Strategy} & \textbf{Qwen3.6-35B} & \textbf{Gemma4-31B} & \textbf{GPT-5.4-mini} \\
\midrule
Query (static)    & 25.9 & 51.6 & 57.6 \\
Single-Pass       & 33.5 & 57.7 & 62.1 \\
Multi-Turn        & 35.2 & 60.2 & 70.7 \\
Scattershot       & 35.4 & 61.2 & 77.6 \\
\textbf{Memetic}  & \textbf{41.9} & \textbf{70.2} & \textbf{84.3} \\
\midrule
$\Delta$ (Memetic\,$-$\,Query) & \textbf{+16.0} & \textbf{+18.6} & \textbf{+26.7} \\
\bottomrule
\end{tabular}
\end{table}

\subsection{RQ1: Pseudo-Tool Descriptions Reduce the Query--Tool Gap}
On \emph{ToolRet} (Code/Web/Customized, Global NDCG), every GPT-5.4-mini reformulation strategy beats the LLM-free Query (static) floor by $+2.7$ to $+10.6$\,pp NDCG@5, and Memetic improves all nine model--domain comparisons in the three-model grid (Table~\ref{tab:main_results_final}). These results show that pseudo-tool descriptions reduce the query--tool gap. Memetic and Scattershot are statistically indistinguishable in the GPT-5.4-mini slice (paired bootstrap, all domains n.s.; Appendix~\ref{app:full_retrieval}) because ToolRet is single-shot: the strategies separate mainly in the end-to-end setting (Table~\ref{tab:toolbench_passrate}).

\subsection{RQ2: Dynamic Retrieval Improves End-to-End Tool Use}
RQ2 asks whether retrieval gains translate into solved tool-use tasks. On StableToolBench, Memetic reaches \textbf{84.3\%} with GPT-5.4-mini: \textbf{$+22.2$ points} over Single-Pass at $62.1\%$ and \textbf{$+26.7$ points} over Query (static) at $57.6\%$ (Table~\ref{tab:toolbench_passrate}). Its gain beyond Single-Pass grows across Qwen3.6-35B, Gemma4-31B, and GPT-5.4-mini ($+8.4\!\to\!+12.5\!\to\!+22.2$), while the Memetic\,$-$\,Query advantage grows $+16.0\!\to\!+18.6\!\to\!+26.7$. This supports a common retrieval interface whose benefit scales with model competence rather than remaining model-invariant. The gains concentrate on the multi-tool G2 subset, where queries are ambiguous and lexical cues are weak: Memetic improves every GPT-5.4-mini split, with the largest jumps on G2-category ($+36.5$) and G2-instruction ($+35.1$; Table~\ref{tab:toolbench_gpt_split}, appendix).

\subsection{Comparisons with Retrieval Baselines}
\label{sec:baselines}
Under a matched retriever and GPT-5.4-mini solver/judge, the accepted-response head-to-head is decisive (Appendix Table~\ref{tab:baselines_h2h}). Memetic reaches $84.3/87.6/82.0\%$ pooled/G2/G3, versus $61.1/68.8/53.9\%$ for \textbf{Re-Invoke} \citep{chen2024re} and $56.8/52.3/56.9\%$ for our \textbf{Xu-style root-refinement} approximation \citep{xu2024enhancing}. Relative to Re-Invoke, this is a $+23.2/+18.8/+28.1$-point gain. Both baselines were reimplemented from their paper descriptions because neither released code; the Xu-style row approximates root-level query interpretation rather than claiming an exact reproduction. Memetic's advantage therefore persists far beyond comparison with static retrieval and is largest in the hard multi-tool regime.

\subsection{RQ3: Dynamic Re-Retrieval Restores Candidate Availability}
\label{sec:recovery}
Table~\ref{tab:frame_widening} isolates the mechanism after a miss: among wrong-first-tool GPT-5.4-mini tasks on G2, dynamic re-retrieval raises top-5 correct-tool availability from $\sim$7--23\% under a frozen root frame to $44$--$83\%$. The first-tool off-gold rate stays within a $\sim$5.5\,pp band across methods, so the gain arises after the initial choice, when revised hypotheses are tested against the corpus.

\begin{table}[H]
\centering
\footnotesize
\caption{Correct-tool availability (\%) among wrong-first-tool GPT-5.4-mini tasks on G2. Values deduplicate the top-5 set across retrievals; static retrieval cannot widen its root frame.}
\label{tab:frame_widening}
\begin{tabular}{@{}lcccc@{}}
\toprule
\textbf{Split} & \textbf{Static} & \textbf{Multi-Turn} & \textbf{Scattershot} & \textbf{Memetic} \\
\midrule
G2-category    & $\sim$7  & 44 & 55 & \textbf{83} \\
G2-instruction & $\sim$23 & 73 & \textbf{80} & 65 \\
\bottomrule
\end{tabular}
\end{table}

\paragraph{Candidate recovery creates the self-correction channel.}
Averaging Multi-Turn, Scattershot, and Memetic, correct-tool availability after a wrong first pick rises from approximately $7\%$ to $61\%$ on G2-category and from $23\%$ to $73\%$ on G2-instruction. Dynamic retrieval therefore restores an action-space opportunity that frozen root retrieval structurally cannot express, explaining why \ourmethod's largest end-to-end gains occur on ambiguous multi-tool tasks.

%% file: mainline_limits.tex
\subsection{RQ4: Efficient Retrieval-Side Test-Time Scaling}
\label{sec:limits}
\paragraph{Performance gain with stable wall-clock.}
Against Single-Pass, Memetic gains $22.2$ pass-rate points ($84.3\%$ vs.\ $62.1\%$). Parallel population execution converts the additional upstream work into nearly unchanged batched wall-clock ($154$ vs.\ $152$ seconds; $1.01\times$), while the downstream solver core remains essentially constant (Appendix Table~\ref{tab:current_latency}).

\paragraph{Single-trajectory refinement plateaus.}
Multi-Turn shows diminishing returns beyond $T{=}1$, motivating population breadth.

\paragraph{Retrieval, not planner scaling, is the binding constraint.}
Across the planner variants evaluated, a stronger planner does not materially improve pass rate. Under the matched GPT-5.4-mini comparison, Xu-style root-only refinement is flat relative to Query ($56.8\%$ vs.\ $57.6\%$), Re-Invoke reaches $61.1\%$, and dynamic Memetic reaches $84.3\%$ (Appendix Table~\ref{tab:baselines_h2h}). Together these results isolate mid-execution retrieval as the component that moves end-to-end performance in this agentic setting.

\paragraph{Benefits scale with model capability.}
The Memetic\,$-$\,Query margin grows $+16.0\!\to\!+18.6\!\to\!+26.7$ across Qwen3.6-35B, Gemma4-31B, and GPT-5.4-mini (Table~\ref{tab:toolbench_passrate}). Thus model-agnostic means a shared interface, not model-invariant performance: stronger semantic variation produces a stronger search.

%% file: conclusion.tex
\ourmethod\ makes retrieval a revisable, corpus-tested search over the agent's action space. Memetic reaches $84.3\%$ pooled pass rate, decisively beating static Query, Re-Invoke, and Xu-style refinement, especially on multi-tool tasks. Planner ablations and recovery identify retrieval as the bottleneck: parallel search restores missing actions, delivering $+22.2$ points over Single-Pass at $1.01\times$ batched wall-clock.

%% file: ai_disclosure.tex
\section*{Use of Generative AI}
OpenAI Codex assisted with manuscript restructuring and with drafting and revising portions of the text. The authors reviewed and verified the final text, citations, analyses, and claims and take responsibility for the paper. Models used as components of the research method and evaluation are identified in the manuscript.

%% file: appendix.tex

\section{Experimental Details}
\label{app:experiments}

\subsection{Ablation Studies}
\label{app:ablation}

Both ablations in this subsection use the \emph{earlier-generation} configuration (GPT-4.1 / Qwen3-30B analysis models); we read the trends qualitatively, as the mechanisms they probe are configuration-independent.

\begin{figure*}[ht]
    \centering
    \includegraphics[width=0.95\textwidth]{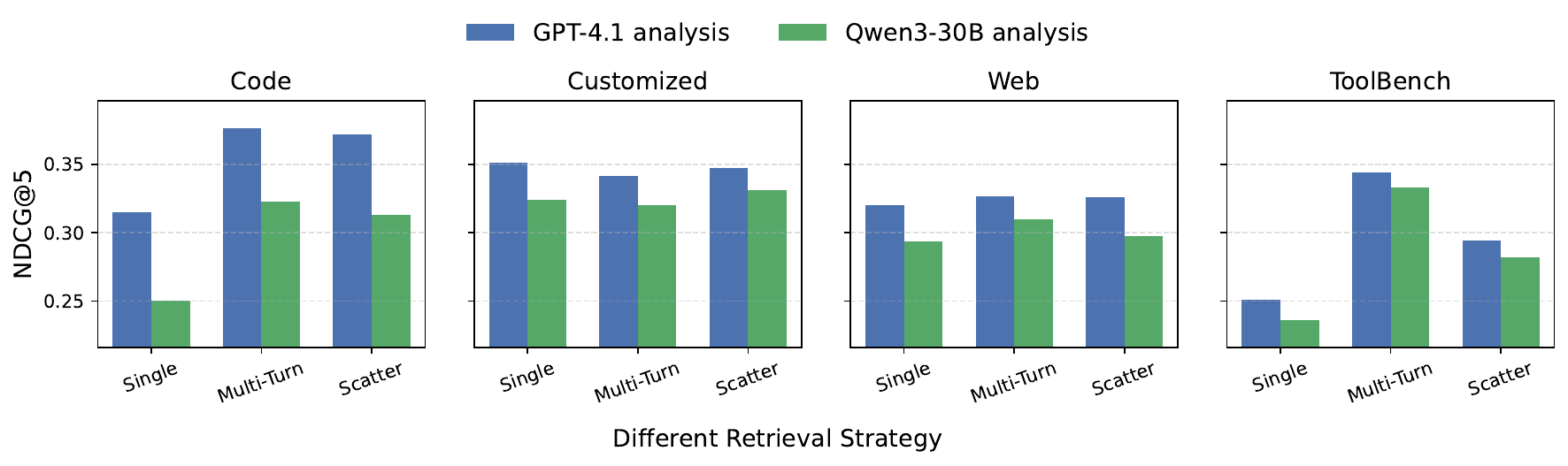}
    \caption{[\textbf{Legacy configuration.}] Effect of query analysis model on retrieval performance. Using GPT-4.1 as the analysis model consistently outperforms Qwen3-30B across all datasets and strategies, indicating that pseudo-tool quality benefits directly from stronger query understanding.}
    \label{fig:ablation_analysis_model}
\end{figure*}

\paragraph{Effect of Query Analysis Model.}
We study how the model used for initial query analysis affects retrieval performance.
As shown in Figure~\ref{fig:ablation_analysis_model}, using a stronger model (GPT-4.1) to interpret the user query leads to consistently higher retrieval scores across all datasets and strategies.
This result suggests that our framework can naturally benefit from better query understanding, as the quality of the initial tool description directly influences subsequent refinement and retrieval. This trend implies that improving the initial task understanding allows even smaller models in later refinement steps to achieve strong performance.

\begin{figure*}[ht]
    \centering
    \includegraphics[width=0.95\textwidth]{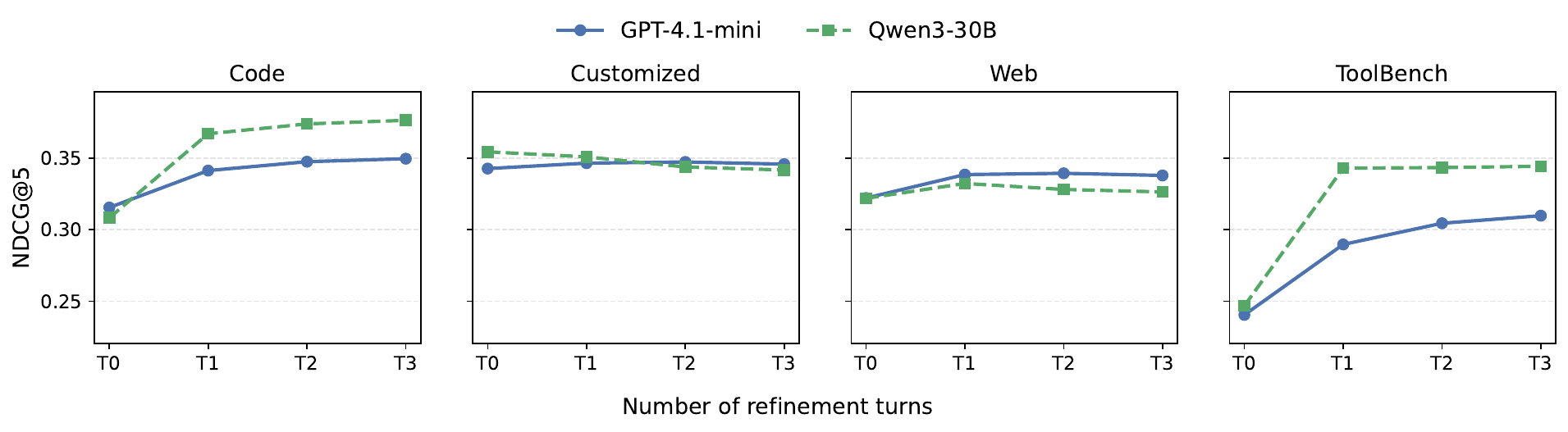}
    \caption{[\textbf{Legacy configuration.}] Effect of refinement turns on retrieval performance. The largest improvement occurs from T0 to T1; subsequent turns yield diminishing returns, suggesting that the first retrieval feedback round provides the most informative grounding signal.}
    \label{fig:ablation_turns}
\end{figure*}

\paragraph{Effect of Number of Refinement Turns.}
As shown in Figure~\ref{fig:ablation_turns}, across most datasets, the improvement from \texttt{T0} to \texttt{T1} is the most significant, while later refinements bring only marginal gains.
This trend aligns with intuition: the first refinement round is when the LLM first receives feedback from the tool database, allowing it to adjust its understanding of which tools actually exist and to optimize its generated descriptions accordingly.
Subsequent iterations mainly provide fine-tuning rather than substantial corrections.
An exception appears in the \emph{Customized} dataset, where performance remains almost unchanged across turns.
This suggests that feedback from retrieved tools may be less informative in this domain, consistent with the observation that \emph{Scattershot} performs better than \emph{Multi-Turn} and that even \emph{Single-Pass} already achieves strong results.

\subsection{Experimental Setup Details}
\label{app:setup}

\paragraph{Tool Corpus Construction.}
The original tool corpus contains detailed metadata for each tool, including parameters, request/response examples, and code snippets. Since our goal is to perform \emph{semantic matching} based on tool descriptions, we simplify these representations by prompting \emph{GPT-4.1-mini} to generate concise natural-language summaries for each tool. Given the full documentation as input, the model produces a short description capturing the tool's core functionality and purpose, removing low-level implementation details while preserving semantic meaning for retrieval.

\paragraph{Tasks and settings.}
We consider two evaluation tasks:
(i) \emph{retrieval-only}: given a single user query, the model retrieves the most relevant tools directly based on the query text without executing or calling any tools; and
(ii) \emph{end-to-end tool use}: the agent performs multi-turn reasoning in an executable environment, actively calling tools and consuming their responses to produce a final answer.

\paragraph{Base models.}
The current experiments use \emph{GPT-5.4-mini}, \emph{Qwen3.6-35B}, and \emph{Gemma4-31B} (\S\ref{sec:setup}). On ToolRet, the named model supplies analysis and refinement through an identical pseudo-tool interface. On StableToolBench, the named suite model is the downstream solver and supplies the reformulation calls for that configuration; GPT-5.4-mini is the judge for every suite.
The \emph{legacy} configuration (Appendix~\ref{app:legacy_extended}) used the closed-source \emph{GPT-4.1-mini} and the open-source \emph{Qwen3-30B} for retrieval-only tasks.
Since Qwen3-30B sometimes underperformed in complex query analysis, consistent with evidence that smaller models struggle with environmental comprehension in agentic settings, relying more heavily on internal reasoning chains~\citep{cuadron2025danger}, that configuration let GPT-4.1 first reformulate the input query, with subsequent refinement (e.g., multi-turn or scattershot strategies) performed by Qwen3-30B.
For legacy end-to-end tool-use tasks, a similar hybrid design applied: \emph{o3-mini} assisted in initial query understanding, and \emph{GPT-4.1-mini} handled downstream reasoning, retrieval, and execution.
Across the planning-model variants evaluated, stronger planning does not materially improve end-to-end pass rate. In contrast, changing the retrieval strategy produces large and consistent gains, supporting retrieval as the binding constraint in this setup.

\paragraph{Metrics.}
We evaluate retrieval performance using multiple standard IR metrics.
\emph{NDCG@5} measures ranking precision with position-aware weighting.
\emph{Comprehensiveness@5} assesses whether all ground-truth tools are successfully retrieved within the top results, reflecting holistic retrieval completeness.
\emph{Precision@5} and \emph{Recall@5} further capture the accuracy and coverage of the retrieved tool set.

\subsection{Benchmark Task Structure}
\label{app:benchmark}

Our end-to-end evaluation uses StableToolBench~\citep{guo2024stabletoolbench}, a stabilized variant of ToolBench~\citep{qintoolllm} that replaces unstable real APIs with a virtual API server combining cached responses and LLM-based API simulators. The benchmark organizes tasks into three complexity levels based on how many tools are required, each further divided by how the test queries were sampled:

\begin{itemize}[leftmargin=*,topsep=2pt,itemsep=2pt,parsep=0pt]
  \item \textbf{G1 (Single-tool):} Tasks solvable with a single API tool. The agent must identify and call one correct tool from the catalog. Three sampling sub-groups:
  \begin{itemize}[topsep=1pt,itemsep=1pt]
    \item \emph{G1-Inst (Instruction)}: Queries derived from human-written task instructions.
    \item \emph{G1-Cat (Category)}: Queries sampled at the tool category level.
    \item \emph{G1-Tool}: Queries sampled at the individual tool level.
  \end{itemize}

  \item \textbf{G2 (Intra-category multi-tool):} Tasks requiring multiple tools from the \emph{same} RapidAPI category. The agent must coordinate calls to related tools (e.g., two weather APIs from the ``Weather'' category). Sub-groups: \emph{G2-Cat}, \emph{G2-Inst}.

  \item \textbf{G3 (Intra-collection multi-tool):} Tasks requiring tools from \emph{different} categories within the same RapidAPI collection. This is the hardest setting: the agent must discover and compose tools across category boundaries. Sub-group: \emph{G3-Inst}.
\end{itemize}

\noindent The progression from G1 to G3 reflects increasing retrieval difficulty: G1 tasks often have strong lexical overlap between the query and the target tool description, while G3 tasks require cross-category reasoning where the semantic gap is widest. This progression explains why advanced retrieval strategies, particularly Memetic, deliver their largest gains on ambiguous multi-tool tasks: simple query-based retrieval is strongest when the task nearly names the tool, whereas composition requires adaptive search.

\subsection{Full End-to-End Results on StableToolBench}
\label{app:full_passrate}

Table~\ref{tab:toolbench_gpt_split} gives the per-split GPT-5.4-mini breakdown summarized in \S\ref{sec:experiments}, and Table~\ref{tab:full_passrate} reports the full earlier-generation strategy grid retained for historical comparison.

\begin{table}[t]
\centering
\small
\caption{GPT-5.4-mini per-split StableToolBench pass rates (\%): static \emph{Query} (SimCSE) vs.\ \emph{Memetic}, with $\Delta$ = Memetic\,$-$\,Query.}
\label{tab:toolbench_gpt_split}
\begin{tabular}{@{}lccc@{}}
\toprule
\textbf{Split} & \textbf{Query} & \textbf{Memetic} & \textbf{$\Delta$} \\
\midrule
G1-category    & 67.6 & 88.2 & +20.6 \\
G1-instruction & 56.0 & 76.7 & +20.7 \\
G1-tool        & 56.0 & 84.2 & +28.2 \\
G2-category    & 53.8 & 90.3 & +36.5 \\
G2-instruction & 49.8 & 84.9 & +35.1 \\
G3-instruction & 60.3 & 82.0 & +21.7 \\
\midrule
\textbf{Pooled} & \textbf{57.6} & \textbf{84.3} & \textbf{+26.7} \\
\bottomrule
\end{tabular}
\end{table}

\textbf{Key observations} (DBD/IBI refinement variants are defined in Appendix~\ref{app:strategies}): (i) DBD with $T{=}3$ outperforms $T{=}5$, showing diminishing returns from additional single-lineage refinement; (ii) IBI varies with the judge configuration, indicating sensitivity to its fitness signal; (iii) Scattershot with $S{=}5$ outperforms $S{=}10$ on most subsets, so diversity saturates quickly; and (iv) Memetic ($N{=}5$, $G{=}3$) achieves the strongest result across every reported GPT-5.4-mini split.

\begin{table*}[t]
\centering
\small
\caption{[\textbf{LEGACY / earlier-generation solver.}] Full pass rate (\%) on StableToolBench across all strategy variants, computed with the earlier-generation \emph{GPT-4.1-mini} solver. All dynamic methods use GPT-4.1-mini with SimCSE-RoBERTa-large embeddings. Best per column in \textbf{bold}. IBI variants were not evaluated on G2/G3 categories due to prohibitively high variance across runs. Current headline results use GPT-5.4-mini / Qwen3.6-35B / Gemma4-31B; these legacy numbers are retained for historical comparability only.}
\label{tab:full_passrate}
\begin{tabular}{llccccccc}
\toprule
\textbf{Family} & \textbf{Variant} & \textbf{G1-C} & \textbf{G1-I} & \textbf{G1-T} & \textbf{G2-C} & \textbf{G2-I} & \textbf{G3-I} & \textbf{Avg.} \\
\midrule
Static & Query Retrieval & 56.4 & 52.6 & 55.6 & 42.3 & 47.8 & 42.1 & 49.47 \\
\midrule
Single-Pass & --- & 61.2 & 59.0 & 59.7 & 55.4 & 56.5 & 50.8 & 57.10 \\
\midrule
\multirow{2}{*}{Multi-Turn (DBD)} & $T{=}3$ & 69.3 & 60.5 & 59.0 & 50.5 & 51.3 & 48.1 & 56.45 \\
 & $T{=}5$ & 67.5 & 60.0 & 61.3 & 55.1 & 51.6 & 32.5 & 54.67 \\
\midrule
\multirow{4}{*}{Multi-Turn (IBI)} & $T{=}3$, $j{=}0$ & 66.0 & 64.8 & 55.8 & --- & --- & --- & 62.20 \\
 & $T{=}3$, $j{=}1$ & 70.6 & 63.0 & 51.4 & --- & --- & --- & 61.67 \\
 & $T{=}5$, $j{=}0$ & 64.5 & 60.3 & 13.7 & --- & --- & --- & 46.17 \\
 & $T{=}5$, $j{=}1$ & 65.1 & 59.9 & 42.5 & --- & --- & --- & 55.83 \\
\midrule
\multirow{2}{*}{Scattershot} & $S{=}5$ & 62.6 & 68.7 & 60.5 & 55.8 & 55.0 & 52.7 & 59.22 \\
 & $S{=}10$ & 64.7 & 58.9 & 60.8 & 61.6 & 56.3 & 47.8 & 58.35 \\
\midrule
\multirow{2}{*}{Memetic} & $P{=}5$, $G{=}3$ & \textbf{77.5} & \textbf{72.4} & \textbf{68.8} & \textbf{70.0} & \textbf{76.4} & \textbf{75.4} & \textbf{73.42} \\
 & V2, $P{=}8$, $G{=}3$ & 61.1 & 62.2 & 64.3 & 58.2 & 57.2 & 56.6 & 59.93 \\
\bottomrule
\end{tabular}
\end{table*}

\begin{table*}[t]
\centering
\small
\caption{[\textbf{LEGACY / capability-threshold diagnostic.}] Pass rate (\%) on StableToolBench using the earlier-generation \textbf{Qwen3-30B} configuration. Its ordering differs from the current suite, in which Memetic leads every model, locating a historical capability threshold for the LLM as a semantic variation operator. Best per column in \textbf{bold}. Current headline results use GPT-5.4-mini / Qwen3.6-35B / Gemma4-31B.}
\label{tab:qwen_passrate}
\begin{tabular}{llccccccc}
\toprule
\textbf{Family} & \textbf{Variant} & \textbf{G1-C} & \textbf{G1-I} & \textbf{G1-T} & \textbf{G2-C} & \textbf{G2-I} & \textbf{G3-I} & \textbf{Avg.} \\
\midrule
Static & Query Retrieval & 47.9 & 30.7 & 32.3 & 41.9 & 39.3 & 16.7 & 34.80 \\
\midrule
Single-Pass & --- & \textbf{58.0} & 46.6 & 41.9 & 47.4 & 45.3 & 35.8 & 45.80 \\
\midrule
\multirow{2}{*}{Multi-Turn (DBD)} & $T{=}3$ & 51.4 & \textbf{49.3} & 39.3 & \textbf{58.2} & 37.7 & \textbf{50.0} & \textbf{47.70} \\
 & $T{=}5$ & 52.2 & 39.5 & 36.0 & 46.4 & 30.7 & 33.6 & 39.70 \\
\midrule
\multirow{2}{*}{Scattershot} & $S{=}5$ & 58.5 & 47.9 & 32.0 & 53.6 & 36.0 & 42.3 & 45.10 \\
 & $S{=}10$ & 49.6 & 36.8 & 40.9 & 50.9 & \textbf{43.7} & 38.0 & 43.30 \\
\midrule
Memetic & $P{=}5$, $G{=}3$ & 42.5 & 31.7 & 27.8 & 38.3 & 36.0 & 16.7 & 32.20 \\
\bottomrule
\end{tabular}
\end{table*}

\paragraph{Model capability as a prerequisite for evolutionary search (legacy solvers).}
Tables~\ref{tab:full_passrate} and~\ref{tab:qwen_passrate} retain \emph{earlier-generation} solvers for historical comparability; the headline suite is GPT-5.4-mini / Qwen3.6-35B / Gemma4-31B. Together, the generations expose a useful scaling law. In the old Qwen3-30B configuration, the shallow DBD $T{=}3$ operating point leads at $47.7\%$, while Memetic reaches $32.2\%$; with legacy GPT-4.1-mini, Memetic leads at $73.4\%$. In the current suite, Memetic leads every model and its margin over Query grows $+16.0\!\to\!+18.6\!\to\!+26.7$ points. The older result therefore locates a capability threshold: once the LLM can supply coherent semantic variation, selection and memory compound useful hypotheses, and stronger models make the search increasingly effective~\citep{cuadron2025danger,novikov2025alphaevolve,guo2025evoprompt,agrawal2025gepa}.

\subsection{Full Retrieval-Only Results on ToolRet}
\label{app:full_retrieval}

\textbf{Memetic vs.\ Scattershot significance.} A paired bootstrap (20k resamples) on GPT-5.4-mini NDCG@5 straddles zero on every domain (Code $+0.66$, Customized $-0.73$, Web $+0.69$; all n.s.), supporting the main-body claim (\S\ref{sec:experiments}) that the two are statistically indistinguishable in the single-shot retrieval-only setting.

Table~\ref{tab:main_results_final} reports the complete domain-level ToolRet retrieval grid summarized in \S\ref{sec:experiments}: all four metrics (NDCG@5, Precision@5, Recall@5, Comprehensiveness@5) for every strategy across all three base models and the three ToolRet domains.

\begin{table*}[t]
\caption{Main results on retrieval-only setting (ToolRet, Global-aggregation NDCG). Each domain reports four metrics: NDCG@5 (N@5), Precision@5 (P@5), Recall@5 (R@5), and Comprehensiveness@5 (C@5). Bold indicates best within each base-model group; \underline{underlined bold} indicates best across all groups. The Query Retrieval floor is LLM-free and identical across models. The Multi-Turn (DBD) row uses $t{=}3$ refinement depth for GPT-5.4-mini and Gemma4-31B; the Qwen3.6-35B Multi-Turn cell uses $t{=}5$.}
\centering
\resizebox{\textwidth}{!}{%
\footnotesize
\setlength{\tabcolsep}{3.2pt}
\renewcommand{\arraystretch}{1.2}
\begin{tabular}{ccccccccccccccc}
\toprule
 \multirow{2}{*}{\textbf{Base Model}}&\multirow{2}{*}{\textbf{Strategy}}  & \multicolumn{4}{c}{\textbf{Code}} & \multicolumn{4}{c}{\textbf{Web}} & \multicolumn{4}{c}{\textbf{Customized}}\\
\cmidrule(lr){3-6}\cmidrule(lr){7-10}\cmidrule(lr){11-14} &
& N@5 & P@5 & R@5 & C@5
& N@5 & P@5 & R@5 & C@5
& N@5 & P@5 & R@5 & C@5 \\
\midrule
--- & Query Retrieval   & 26.85 & 8.74 & 32.16 & 29.33  & 30.06 & 10.46 & 35.57 & 26.73  & 29.87 & 13.14 & 32.25 & 20.77 \\
\midrule
\multirow{4}{*}{\textbf{GPT-5.4-mini}}
& Single-Pass      & 33.01 & 10.01 & 39.13 & 35.62  & 33.18 & 12.32 & 39.95 & 30.36  & 35.02 & 15.72 & 37.93 & 24.95 \\
& Multi-Turn        & 37.06 & \textbf{11.58} & \textbf{43.90} & \textbf{40.31}  & 32.80 & 12.01 & 39.13 & 29.90  & 35.34 & 15.17 & 38.25 & 25.56 \\
& Scattershot       & 36.77 & 11.16 & 43.24 & 39.56  & 34.14 & \textbf{12.71} & \underline{\textbf{40.51}} & \underline{\textbf{30.84}}  & \underline{\textbf{36.63}} & \textbf{16.33} & \underline{\textbf{39.62}} & 25.97 \\
& Memetic            & \textbf{37.43} & 11.04 & 42.30 & 39.56  & \textbf{34.83} & 11.88 & 39.18 & 28.68  & 35.90 & 14.05 & 37.52 & \underline{\textbf{26.38}} \\
\midrule
\multirow{4}{*}{\textbf{Qwen3.6-35B}}
& Single-Pass      & 30.12 & 9.74 & 37.74 & 34.08  & 30.92 & 11.38 & 37.26 & 28.38  & 32.66 & 14.99 & 35.82 & 23.22 \\
& Multi-Turn        & 38.17 & 11.85 & 44.05 & 40.02  & \textbf{34.44} & \underline{\textbf{12.89}} & \textbf{40.46} & \textbf{30.78}  & 33.89 & \textbf{15.32} & 35.58 & 22.61 \\
& Scattershot       & 34.58 & 11.22 & 42.42 & 37.96  & 31.99 & 12.03 & 38.98 & 29.71  & 33.20 & 14.38 & 36.21 & 24.03 \\
& Memetic            & \textbf{39.42} & \textbf{11.93} & \underline{\textbf{45.03}} & \textbf{41.51}  & 34.41 & 11.67 & 39.15 & 29.04  & \textbf{35.09} & 14.09 & \textbf{36.56} & \textbf{25.25} \\
\midrule
\multirow{4}{*}{\textbf{Gemma4-31B}}
& Single-Pass      & 25.95 & 8.95 & 33.08 & 28.87  & 30.96 & 11.39 & 37.55 & 29.18  & 34.18 & 15.91 & 37.67 & 24.44 \\
& Multi-Turn        & 33.69 & 11.22 & 39.16 & 36.59  & 32.81 & 10.79 & 36.77 & 26.75  & 32.76 & 11.98 & 33.81 & 24.24 \\
& Scattershot       & 31.22 & 10.67 & 37.30 & 33.33  & 33.59 & \textbf{12.55} & \textbf{40.33} & \textbf{30.69}  & \textbf{36.25} & \underline{\textbf{16.62}} & \textbf{39.61} & \textbf{25.97} \\
& Memetic            & \underline{\textbf{39.73}} & \underline{\textbf{12.10}} & \textbf{44.82} & \underline{\textbf{41.68}}  & \underline{\textbf{35.35}} & 11.76 & 39.49 & 28.89  & 34.59 & 13.46 & 35.51 & 24.34 \\
\bottomrule
\end{tabular}}
\label{tab:main_results_final}
\end{table*}

\subsection{Head-to-Head Baseline Comparison}
\label{app:baselines}
Table~\ref{tab:baselines_h2h} reports the full head-to-head against retrieval baselines summarized in \S\ref{sec:baselines}.
\begin{table}[ht]
\centering
\small
\caption{Retrieval-baseline comparisons. \emph{Top:} accepted-response StableToolBench end-to-end head-to-head under a matched \texttt{sup-simcse-roberta-large} retriever and GPT-5.4-mini solver/judge. Both prior methods were reimplemented from their papers because code was unavailable; the Xu-style row approximates root-level query interpretation. \emph{Bottom:} secondary ToolRet retrieval-only NDCG@5 diagnostic by domain; the one-lineage Multi-Turn row is not an exact Xu et al.\ reproduction.}
\label{tab:baselines_h2h}
\begin{tabular}{@{}lrrr@{}}
\toprule
\multicolumn{4}{l}{\emph{StableToolBench end-to-end pass rate (\%)}} \\
\textbf{Method} & \textbf{Pooled} & \textbf{G2} & \textbf{G3} \\
\midrule
Query (static)                       & 57.6 & 51.8 & 60.3 \\
Xu-style root refinement \citep{xu2024enhancing} & 56.8 & 52.3 & 56.9 \\
Re-Invoke \citep{chen2024re}         & 61.1 & 68.8 & 53.9 \\
\textbf{Memetic (ours)}              & \textbf{84.3} & \textbf{87.6} & \textbf{82.0} \\
\midrule
$\Delta$ (Memetic\,$-$\,Query)     & \textbf{+26.7} & \textbf{+35.8} & \textbf{+21.7} \\
\bottomrule
\end{tabular}

\vspace{4pt}
\begin{tabular}{@{}lccc@{}}
\toprule
\multicolumn{4}{l}{\emph{ToolRet retrieval-only NDCG@5 (Global)}} \\
\textbf{Method} & \textbf{Code} & \textbf{Web} & \textbf{Customized} \\
\midrule
Query (SimCSE, static)              & 26.85 & 30.06 & 29.87 \\
Re-Invoke \citep{chen2024re}       & 31.35 & 34.64 & 39.21 \\
Multi-Turn (one-lineage query-side diagnostic) & 37.06 & 32.80 & 35.34 \\
\textbf{Memetic (ours)}            & \textbf{37.43} & \textbf{34.83} & 35.90 \\
\bottomrule
\end{tabular}
\end{table}

\subsection{Per-Subset Retrieval Results for Memetic}
\label{app:per_subset}

Tables~\ref{tab:memetic_code}--\ref{tab:memetic_web} report per-subset retrieval metrics for \emph{Memetic} across the three ToolRet domains where results are available.
These were obtained with the \emph{legacy} GPT-4.1-mini configuration (Appendix~\ref{app:legacy_gpt41}) and provide subset-level granularity; the current-model domain-level grid is Table~\ref{tab:main_results_final}.

\begin{table}[h]
\centering
\small
\caption{[\textbf{Legacy GPT-4.1-mini.}] Memetic per-subset results on ToolRet \textbf{Code} domain. Global metrics use merged qrels.}
\label{tab:memetic_code}
\begin{tabular}{lrcccccccc}
\toprule
& & \multicolumn{4}{c}{\textbf{GPT-4.1-mini}} & \multicolumn{4}{c}{\textbf{Qwen3-30B}} \\
\cmidrule(lr){3-6}\cmidrule(lr){7-10}
\textbf{Subset} & \textbf{\#Q} & N@5 & P@5 & R@5 & C@5 & N@5 & P@5 & R@5 & C@5 \\
\midrule
gorilla-pytorch     & 43  & 11.14 & 3.72 & 18.61 & 18.61 & 8.44 & 1.86 & 9.30 & 9.30 \\
gorilla-tensor      & 55  & 3.10 & 1.45 & 7.27 & 7.27 & 3.31 & 1.09 & 5.45 & 5.45 \\
gorilla-huggingface & 500 & 24.15 & 6.52 & 32.60 & 32.60 & 23.23 & 6.96 & 34.80 & 34.80 \\
craft-tabmwp        & 174 & 12.92 & 3.79 & 18.97 & 18.97 & 15.19 & 4.14 & 20.69 & 20.69 \\
craft-vqa           & 200 & 11.52 & 2.90 & 14.50 & 14.50 & 12.51 & 2.90 & 14.50 & 14.50 \\
craft-math-algebra  & 280 & 64.29 & 14.64 & 73.21 & 73.21 & 60.31 & 14.07 & 70.36 & 70.36 \\
toolink             & 497 & 52.80 & 21.05 & 54.52 & 41.25 & 54.23 & 20.81 & 55.64 & 41.45 \\
\midrule
\textbf{Global}     & 1749 & \textbf{35.18} & \textbf{11.04} & \textbf{40.76} & \textbf{36.99} & \textbf{34.96} & \textbf{10.98} & \textbf{41.14} & \textbf{37.11} \\
\bottomrule
\end{tabular}
\end{table}

\begin{table}[h]
\centering
\small
\caption{[\textbf{Legacy GPT-4.1-mini.}] Memetic per-subset results on ToolRet \textbf{Customized} domain.}
\label{tab:memetic_customized}
\begin{tabular}{lrcccccccc}
\toprule
& & \multicolumn{4}{c}{\textbf{GPT-4.1-mini}} & \multicolumn{4}{c}{\textbf{Qwen3-30B}} \\
\cmidrule(lr){3-6}\cmidrule(lr){7-10}
\textbf{Subset} & \textbf{\#Q} & N@5 & P@5 & R@5 & C@5 & N@5 & P@5 & R@5 & C@5 \\
\midrule
gpt4tools           & 32  & 60.32 & 14.38 & 71.88 & 71.88 & 55.87 & 14.38 & 71.88 & 71.88 \\
taskbench-hf        & 23  & 42.08 & 21.74 & 47.07 & 30.43 & 26.87 & 16.52 & 37.46 & 21.74 \\
taskbench-mm        & 40  & 73.07 & 34.00 & 75.02 & 55.00 & 65.72 & 32.00 & 72.33 & 52.50 \\
toolbench-sam       & 197 & 7.49 & 1.93 & 9.65 & 9.65 & 8.32 & 2.44 & 12.18 & 12.18 \\
toolalpaca          & 94  & 54.80 & 13.62 & 68.09 & 68.09 & 59.41 & 14.47 & 72.34 & 72.34 \\
gta                 & 14  & 13.14 & 4.29 & 10.71 & 0.00 & 29.04 & 8.57 & 25.00 & 7.14 \\
tool-be-honest      & 350 & 35.88 & 25.94 & 31.11 & 0.00 & 33.05 & 25.14 & 29.67 & 0.00 \\
appbench            & 32  & 75.64 & 32.50 & 82.55 & 59.38 & 71.45 & 28.75 & 72.14 & 50.00 \\
metatool            & 200 & 37.47 & 9.30 & 44.50 & 42.50 & 29.60 & 7.50 & 35.50 & 34.50 \\
\midrule
\textbf{Global}     & 982 & \textbf{35.75} & \textbf{16.31} & \textbf{37.95} & \textbf{24.34} & \textbf{33.03} & \textbf{15.58} & \textbf{36.05} & \textbf{23.12} \\
\bottomrule
\end{tabular}
\end{table}

\begin{table*}[h]
\centering
\small
\caption{[\textbf{Legacy GPT-4.1-mini.}] Memetic per-subset results on ToolRet \textbf{Web} domain (19 subsets, 5{,}230 queries).}
\label{tab:memetic_web}
\resizebox{\textwidth}{!}{%
\begin{tabular}{lrcccccccc}
\toprule
& & \multicolumn{4}{c}{\textbf{GPT-4.1-mini}} & \multicolumn{4}{c}{\textbf{Qwen3-30B}} \\
\cmidrule(lr){3-6}\cmidrule(lr){7-10}
\textbf{Subset} & \textbf{\#Q} & N@5 & P@5 & R@5 & C@5 & N@5 & P@5 & R@5 & C@5 \\
\midrule
autotools-weather   & 11  & 1.33 & 1.82 & 1.82 & 0.00 & 4.47 & 3.64 & 5.30 & 0.00 \\
autotools-food      & 22  & 33.25 & 14.55 & 34.85 & 13.64 & 32.18 & 18.18 & 43.18 & 18.18 \\
autotools-music     & 32  & 8.35 & 7.50 & 9.32 & 0.00 & 6.40 & 5.00 & 6.09 & 0.00 \\
restgpt-spotify     & 40  & 24.37 & 10.00 & 24.58 & 15.00 & 26.99 & 11.50 & 25.63 & 12.50 \\
restgpt-tmdb        & 54  & 5.93 & 2.96 & 6.48 & 0.00 & 12.31 & 4.81 & 11.11 & 0.00 \\
toolemu             & 38  & 11.20 & 3.68 & 15.79 & 13.16 & 12.14 & 3.16 & 13.16 & 10.53 \\
mnms                & 33  & 25.28 & 12.73 & 26.77 & 6.06 & 16.38 & 9.70 & 19.19 & 6.06 \\
taskbench-daily     & 40  & 27.01 & 11.50 & 27.75 & 15.00 & 11.98 & 6.50 & 17.04 & 12.50 \\
t-eval-dialog       & 50  & 30.58 & 8.40 & 42.00 & 42.00 & 27.70 & 8.80 & 44.00 & 44.00 \\
t-eval-step         & 50  & 35.64 & 21.60 & 34.82 & 8.00 & 31.22 & 19.60 & 31.71 & 8.00 \\
tooleyes            & 95  & 16.57 & 4.63 & 23.16 & 23.16 & 17.44 & 4.84 & 24.21 & 24.21 \\
apibank             & 101 & 23.51 & 7.72 & 26.73 & 16.83 & 24.34 & 7.33 & 26.65 & 18.81 \\
reversechain        & 200 & 19.90 & 11.40 & 19.88 & 2.50 & 13.94 & 7.80 & 13.77 & 1.00 \\
toollens            & 314 & 3.95 & 2.23 & 4.83 & 1.27 & 4.78 & 2.80 & 6.10 & 1.59 \\
rotbench            & 550 & 4.99 & 1.75 & 8.73 & 8.73 & 5.13 & 1.89 & 9.36 & 9.27 \\
ultratool           & 500 & 46.78 & 19.88 & 50.89 & 31.20 & 40.70 & 17.72 & 45.62 & 27.40 \\
apigen              & 1000 & 44.28 & 13.32 & 54.55 & 49.60 & 42.84 & 13.02 & 53.08 & 49.10 \\
toolace             & 1000 & 54.20 & 16.06 & 62.61 & 58.30 & 48.46 & 14.52 & 57.98 & 54.30 \\
toolbench           & 1100 & 32.05 & 15.71 & 35.48 & 17.09 & 31.51 & 15.53 & 35.12 & 16.45 \\
\midrule
\textbf{Global}     & 5230 & \textbf{33.84} & \textbf{12.51} & \textbf{39.33} & \textbf{29.94} & \textbf{31.49} & \textbf{11.78} & \textbf{37.43} & \textbf{28.64} \\
\bottomrule
\end{tabular}}
\end{table*}

\section{Method Details}
\label{app:method_details}

\subsection{DFSDT Integration Details}
\label{app:dfsdt}

All retrieval strategies in \ourmethod operate within the Depth-First Search Decision Tree (DFSDT) reasoning loop \citep{qintoolllm}. Pseudo-tool generation hooks into the DFSDT reasoning chain as follows: when the LLM generates a thought node containing pseudo-tool description blocks, the system triggers a dynamic retrieval call. The choice of retrieval strategy (Single-Pass, IBI, DBD, Scattershot, or Memetic) is determined by the configured policy. Retrieved tools are injected into the active function schema for subsequent action nodes.

\paragraph{Notation.}
Let $\mathcal{N}$ denote the set of tree nodes, where each node $n$ carries a message history $\mathcal{H}(n)$, type $\tau(n) \in \{\text{Thought, Action, ActionInput}\}$, depth $\delta(n)$, and child set $\mathrm{ch}(n)$. Let $\mathcal{F}(n)$ be the function schema (tool list) active at node $n$, and let $\textsc{Exec}(a, x, \mathcal{F})$ execute action $a$ with arguments $x$ against the API environment, returning an observation and status code $\kappa \in \{0{=}\text{ok},\; 1{=}\text{halluc},\; 2{=}\text{error},\; 3{=}\text{answer},\; 4{=}\text{give\_up}\}$. We write $D_{\max}$ for the maximum branch depth, $W$ for beam width, $Q_{\max}$ for the LLM call budget, and $\ell_{\mathrm{prune}}, \ell_{\mathrm{ans}}$ for backtrack distances.

\begin{algorithm}[ht]
\caption{DFSDT Reasoning Loop with Dynamic Retrieval}
\label{alg:dfsdt}
\footnotesize
\begin{algorithmic}[1]
\Require Node $n$, $D_{\max}$, $W$, $Q_{\max}$, answer set $\mathcal{S}$, memory $\Mem$, function schema $\mathcal{F}$
\Ensure Backtrack distance $b$; modifies $\mathcal{S}$, $\Mem$, $\mathcal{F}$
\If{$\delta(n) \geq D_{\max}$ or $n.\mathrm{pruned}$}
  \State \Return $\ell_{\mathrm{prune}}$
\EndIf
\If{$n.\mathrm{terminal}$}
  \State $\mathcal{S}.\mathrm{add}(n)$
  \State \Return $\ell_{\mathrm{ans}}$
\EndIf
\For{$i \gets 1$ to $W$}
  \If{$|\mathcal{S}| \geq \mathcal{S}_{\max}$ or $Q > Q_{\max}$} \Return $\infty$ \EndIf
  \If{$\mathrm{ch}(n) \neq \emptyset$}
    \State Inject diversity prompt summarising prior children
  \EndIf
  \State $o \gets \mathrm{LLM}(\mathcal{H}(n), \mathcal{F}(n))$; \enspace $Q \gets Q + 1$
  \If{$o$ contains pseudo-tool blocks}
    \State $n_{\mathrm{th}} \gets$ Thought child
    \If{not duplicate in $\Mem$}
      \State $\mathit{tools} \gets \textsc{SelectAndRunStrategy}(o, \mathcal{F}, \mathrm{LLM}, \Mem)$
      \State $\mathcal{F}(n_{\mathrm{th}}) \gets \mathcal{F}(n_{\mathrm{th}}) \cup \mathit{tools}$
      \State $\Mem.\mathrm{update}(o)$
    \EndIf
  \EndIf
  \For{each function call $(a, x) \in o$}
    \State $n_a \gets$ Action child; \enspace $n_x \gets$ ActionInput child
    \State $(\mathrm{obs}, \kappa) \gets \textsc{Exec}(a, x, \mathcal{F}(n))$
    \If{$\kappa = 1$} replace $a$ with sentinel \EndIf
    \If{$\kappa = 3$} $n_x.\mathrm{terminal} \gets \mathrm{True}$ \EndIf
    \If{$\kappa \in \{1, 2, 4\}$} $n_x.\mathrm{pruned} \gets \mathrm{True}$ \EndIf
  \EndFor
  \For{each child $c \in \mathrm{ch}(n)$}
    \State $b \gets \textsc{DFSDT}(c, D_{\max}, W, Q_{\max}, \mathcal{S}, \Mem, \mathcal{F})$
    \If{$|\mathcal{S}| \geq \mathcal{S}_{\max}$} \Return $\infty$ \EndIf
    \If{$b > 1$} \Return $b - 1$ \EndIf
  \EndFor
\EndFor
\State \Return $1$
\end{algorithmic}
\end{algorithm}

\paragraph{Mathematical formalization.}
Let the search tree be $\mathcal{G} = (\mathcal{N}, E)$ where each edge $(n, c) \in E$ corresponds to one LLM generation step. (We use $\mathcal{G}$ to avoid overloading $\Corpus$ which denotes the tool corpus.) The message history at node $n$ along path $\pi(n) = (n_0, n_1, \ldots, n)$ is:
\begin{equation}
  \mathcal{H}(n) = \bigl[\mathrm{sys}(q)\bigr] \circ \bigl[\mathrm{usr}(q)\bigr] \circ \bigl(a_i, x_i, o_i\bigr)_{i \in \pi(n)},
\end{equation}
where $(a_i, x_i, o_i)$ are action/argument/observation triples. Backtracking propagates when $b > 1$, continuing expansion only when $b \leq 1$. Dynamic retrieval triggers when:
\begin{equation}
  \max_{m \in \Mem} \mathrm{SequenceSim}(o, m) < 0.82 \;\wedge\; o \not\subset m \;\;\forall m \in \Mem.
\end{equation}

\subsection{Strategy Details: Scattershot, DBD, and IBI}
\label{app:strategies}

We detail the three non-evolutionary strategies below. \textbf{Naming:} the main body's \emph{Multi-Turn Refinement} (\S\ref{sec:method-multiturn}) is implemented as DBD (description-by-description, per-lineage refinement); IBI (iterative bootstrapped improvement) is an alternative global-refinement variant evaluated only with legacy solvers (Appendix~\ref{app:legacy_extended}). All use the shared notation: $q$ (query), $\Corpus$ (tool corpus), $\mathbf{e}(\cdot)$ (sentence embedding via SimCSE-RoBERTa-large), $\Ret(d, \Corpus, k)$ (top-$k$ retrieval by cosine similarity), and $\mathrm{blurb}(t)$ (normalized natural-language description of tool $t$).

\paragraph{DBD (Description-by-Description).}
DBD maintains independent per-lineage refinement paths, each anchored to its original ancestor description. Unlike IBI, exemplars are lineage-local and the ancestor is included in every refinement prompt as an anchor.

\begin{algorithm}[ht]
\caption{DBD --- Per-Lineage Iterative Refinement}
\label{alg:dbd}
\footnotesize
\begin{algorithmic}[1]
\Require LLM output $o$, corpus $\Corpus$, top-$k$, turns $n_t$
\Ensure API key list
\State $\mathcal{D}_0 \gets \textsc{ParseBlocks}(o)$
\For{each ancestor $a \in \mathcal{D}_0$}
  \State $d \gets a$; \enspace $\mathcal{E} \gets \emptyset$
  \For{$t \gets 1$ to $n_t$}
    \State $(Q, R) \gets \Ret(d, \Corpus, k)$
    \State $\mathcal{E} \gets \mathcal{E} \cup \{\mathrm{blurb}(r) : r \in R\}$
    \State $d' \gets \textsc{ParseSingle}(\mathrm{LLM}_{\mathrm{refine}}(a, d, \mathcal{E}, q))$
    \If{$d' \neq \emptyset$} $d \gets d'$ \EndIf
  \EndFor
  \State Append $d$ to finals
\EndFor
\State \Return $\bigcup_{d \in \mathit{finals}} \Ret(d, \Corpus, k).\mathrm{api\_list}$
\end{algorithmic}
\end{algorithm}

For each ancestor $a \in \mathcal{D}_0$, define $d_a^{(0)} = a$ and $\mathcal{E}_a^{(0)} = \emptyset$. At turn $t$:
\begin{align}
  \mathcal{E}_a^{(t)} &= \mathcal{E}_a^{(t-1)} \cup \{\mathrm{blurb}(r) : r \in \Ret(d_a^{(t-1)}, \Corpus, k)\}, \\
  d_a^{(t)} &= \textsc{ParseSingle}\!\left(\mathrm{LLM}_{\mathrm{refine}}(a, d_a^{(t-1)}, \mathcal{E}_a^{(t)}, q)\right).
\end{align}
Final output: $\mathrm{Output} = \bigcup_{a \in \mathcal{D}_0} \Ret(d_a^{(n_t)}, \Corpus, k)$.
LLM cost per lineage: $n_t$ calls; retrievals: $n_t + 1$.

\paragraph{IBI (Iterative Bootstrapped Improvement).}
IBI applies a single global refinement prompt across all descriptions simultaneously. It accumulates exemplar tool blurbs across iterations and optionally uses an LLM judge to check sufficiency.

\begin{algorithm}[ht]
\caption{IBI --- Iterative Bootstrapped Improvement}
\label{alg:ibi}
\footnotesize
\begin{algorithmic}[1]
\Require LLM output $o$, corpus $\Corpus$, top-$k$, turns $n_t$, sufficiency flag \textsc{check}
\Ensure API key list
\State $\mathcal{D} \gets \textsc{ParseBlocks}(o)$; \enspace $\mathcal{B} \gets \emptyset$
\For{$t \gets 1$ to $n_t$}
  \For{each $d \in \mathcal{D}$}
    \State $(Q, R) \gets \Ret(d, \Corpus, k)$
    \State $\mathcal{B} \gets \mathcal{B} \cup \{\mathrm{blurb}(r) : r \in R\}$
  \EndFor
  \If{\textsc{check} and $\mathrm{LLM}_{\mathrm{judge}}(q, \mathcal{B}) = \mathrm{True}$}
    \State \textbf{break}
  \EndIf
  \State $o' \gets \mathrm{LLM}_{\mathrm{refine}}(\mathcal{D}, \mathcal{B}, q)$
  \State $\mathcal{D}' \gets \textsc{ParseBlocks}(o')$
  \If{$\mathcal{D}' = \emptyset$} \textbf{break} \EndIf
  \State $\mathcal{D} \gets \mathcal{D}'$
\EndFor
\For{each $d \in \mathcal{D}$}
  \State Append $\Ret(d, \Corpus, k).\mathrm{api\_list}$ to keys
\EndFor
\State \Return $\mathit{keys}$
\end{algorithmic}
\end{algorithm}

Let $\mathcal{D}^{(0)}$ be the initial description set. At each turn $t$:
\begin{align}
  \mathcal{B}^{(t)} &= \mathcal{B}^{(t-1)} \cup \bigcup_{d \in \mathcal{D}^{(t-1)}} \{\mathrm{blurb}(r) : r \in \Ret(d, \Corpus, k)\}, \\
  \mathcal{D}^{(t)} &= \textsc{ParseBlocks}\!\left(\mathrm{LLM}_{\mathrm{refine}}(\mathcal{D}^{(t-1)}, \mathcal{B}^{(t)}, q)\right).
\end{align}
LLM cost (worst case, $n_t$ turns, $m$ descriptions): $n_t$ refinement calls $+ n_t$ optional judge calls; retrievals: $(n_t + 1) \cdot m$.

\paragraph{Stability: why DBD over IBI.}
The choice between IBI (global refinement) and DBD (per-lineage refinement) has significant implications for stability. IBI lacks pseudo-tool provenance tracking: at each iteration a variable number of descriptions may be generated, making it difficult to attribute improvements to specific lineages. DBD maintains independent refinement paths with explicit lineage anchors, yielding more stable convergence; this seemingly minor design choice proved critical in practice, which is why the main body's Multi-Turn Refinement is the DBD instantiation.

\paragraph{Scattershot.}
Scattershot introduces population-level diversity by sampling multiple candidate descriptions per ancestor in parallel at high temperature, then combining via population-level voting.

\begin{algorithm}[ht]
\caption{Scattershot --- Parallel Diversity Sampling}
\label{alg:scattershot}
\footnotesize
\begin{algorithmic}[1]
\Require LLM output $o$, corpus $\Corpus$, top-$k$, fan-out size $S$, temperature $T_b$, budget $B$
\Ensure API key list
\State $\mathcal{D}_0 \gets \textsc{ParseBlocks}(o)$
\For{each ancestor $a \in \mathcal{D}_0$}
  \State $(Q_0, R_0) \gets \Ret(a, \Corpus, k)$
  \State $\mathcal{E} \gets \{\mathrm{blurb}(r) : r \in R_0\}$
  \State $C \gets \textsc{Parallel}\!\left[\mathrm{LLM}_{\mathrm{scatter}}(a, \mathcal{E};\; T_b)\right]_{i=1}^{S}$
  \State $\Pop \gets \{c_i : c_i \gets \textsc{ParseSingle}(C_i),\; c_i \neq \emptyset\}$
  \State Extend keys with $\textsc{Vote}(\Pop, \Corpus, k, B)$
\EndFor
\State \Return $\mathit{keys}$
\end{algorithmic}
\end{algorithm}

The following subroutines are shared between Scattershot and Memetic (Algorithm~\ref{alg:memetic}):

\begin{algorithm}[ht]
\caption{Seed Population}\label{alg:seedpop}
\footnotesize
\begin{algorithmic}[1]
\Require Ancestor $a$, $\Corpus$, $k$, $N$, $\tau$
\Ensure Population $\Pop$
\State $R_0 \gets \Ret(a, \Corpus, k)$
\State $\mathcal{E} \gets \{\mathrm{blurb}(r) : r \in R_0\}$
\State $\Pop \gets \LLM_{\mathrm{seed}}(a, \mathcal{E}, N;\; \tau)$
\State \Return $\Pop$
\end{algorithmic}
\end{algorithm}

\begin{algorithm}[ht]
\caption{Hierarchical Voting}\label{alg:vote}
\footnotesize
\begin{algorithmic}[1]
\Require Population $\Pop$, $\Corpus$, $k$, budget $B$
\Ensure Top-$B$ ranked tool list
\For{each $d_i \in \Pop$}
  \State $R_i \gets \Ret(d_i, \Corpus, k)$
\EndFor
\State $\mathcal{C} \gets \bigcup_i R_i$ \hfill\emph{candidate pool}
\For{each tool $t \in \mathcal{C}$}
  \State $v(t) \gets |\{i : t \in R_i\}|$ \hfill\emph{vote count}
  \State $\bar{r}(t) \gets \mathrm{mean}_{i: t \in R_i} \mathrm{rank}_i(t)$
  \State $\bar{s}(t) \gets \mathrm{mean}_{i: t \in R_i} \mathrm{sim}_i(t)$
\EndFor
\State Sort $\mathcal{C}$ by $(v\!\downarrow,\; \bar{r}\!\uparrow,\; \bar{s}\!\downarrow)$
\State \Return $\mathrm{top}_B(\mathcal{C})$
\end{algorithmic}
\end{algorithm}

For each ancestor $a$, let $\mathcal{E}_a = \{\mathrm{blurb}(r) : r \in \Ret(a, \Corpus, k)\}$ be seed exemplars. Draw a population of $S$ children independently:
\begin{equation}
  d_i \sim \mathrm{LLM}_{\mathrm{scatter}}(a, \mathcal{E}_a;\; T_b), \quad i = 1, \ldots, S,
\end{equation}
where $T_b = 1.5$ is fixed. For each ancestor $a$, $\textsc{Vote}$ is applied to $\{d_1, \ldots, d_S\}$.
Scattershot is a boundary case of the common design space underlying all our strategies: it corresponds to Memetic with the evolutionary operators disabled ($G{=}1$, no selection, no fitness evaluation, and no local search), retaining only diverse generation and population voting.
LLM cost per lineage: $S$ parallel calls; retrievals: $S + 1$.

\subsection{Hyperparameter Settings and Notation}
\label{app:hyperparams}

Table~\ref{tab:hyperparams} lists key hyperparameters. Table~\ref{tab:notation} summarizes the mathematical notation.

\begin{table}[ht]
\centering
\renewcommand{\arraystretch}{1.15}
\small
\caption{Hyperparameter settings. Search-budget and fitness parameters apply to all experiments; the analysis/refinement model rows describe the \emph{legacy} configuration (current headline base models in \S\ref{sec:setup}).}
\label{tab:hyperparams}
\begin{tabular}{@{}llc@{}}
\toprule
\textbf{Parameter} & \textbf{Description} & \textbf{Value} \\
\midrule
$k$ & Top-$k$ retrieval candidates & 5 \\
$\mathbf{e}(\cdot)$ & Embedding model & SimCSE-RoBERTa-large \\
Analysis LLM & Query analysis model (legacy config.) & GPT-4.1 \\
Refinement LLM & Description refinement model (legacy config.) & GPT-4.1-mini / Qwen3-30B \\
\midrule
$T$ & Multi-Turn refinement turns & 3 or 5 \\
$\tau_{\mathrm{refine}}$ & Multi-Turn temperature & $\leq 0.7$ \\
$S$ & Scattershot fan-out size & 5 or 10 \\
$\tau_{\mathrm{scatter}}$ & Scattershot generation temperature & 1.5 \\
$N$ & Memetic population size & 5 \\
$G$ & Memetic generations & 3 \\
$\theta$ & Memetic retrieval-confidence threshold $r(d^*)$ & 0.95 \\
$\tau_{\mathrm{memetic}}$ & Memetic base temperature & 1.5 \\
$p_{\mathrm{cross}}$ & Crossover branch probability & 0.5 \\
$\alpha$ & Top-1 weight in $r(d)$ (Eq.~\ref{eq:retrieval-confidence}) & 0.7 \\
$m$ & Observation size, top-$m$ retrieval mass (Eq.~\ref{eq:fitness}) & 3 \\
$\lambda$ & KL-drift penalty weight (Eq.~\ref{eq:fitness}) & 1.0 \\
$\sigma$ & KDE kernel bandwidth (Eq.~\ref{eq:fitness}) & 0.5 \\
$B$ & Voting budget (top-$B$ output) & 5 \\
\midrule
\multicolumn{3}{@{}l}{\textit{Tool corpus sizes}} \\
\midrule
StableToolBench & Total tool corpus & 16,464 APIs \\
ToolRet & Total tool corpus & 43k tools \\
\bottomrule
\end{tabular}
\renewcommand{\arraystretch}{1.0}
\end{table}

\begin{table}[ht]
\centering
\small
\caption{Summary of mathematical notation.}
\label{tab:notation}
\begin{tabular}{@{}ll@{}}
\toprule
\textbf{Symbol} & \textbf{Definition} \\
\midrule
$q$ & User query \\
$a_t \in \mathcal{A}$ & Action at time step $t$ \\
$\mathcal{A}$ & Action space \\
$\mathcal{F}_t \subseteq \Corpus$ & Function schema (active tools) at step $t$ \\
$\Corpus = \{t_1, \ldots, t_n\}$ & Tool corpus \\
$d$ & Pseudo-tool description \\
$\mathcal{D}_0$ & Initial pseudo-tool descriptions (ancestors) \\
$\mathcal{H}$ & Message history \\
$\mathbf{e}(\cdot) \to \mathbb{R}^d$ & Sentence embedding function \\
$R \subseteq \Corpus$ & Retrieved tool set (top-$k$) \\
$\mathcal{E}$ & Exemplar set (tool blurbs) \\
$r(d_i) \in [0,1]$ & Unpenalized retrieval-confidence score \\
$f(d_i) \in \mathbb{R}$ & Memory-penalized ranking fitness \\
$\pi_d$ & Retrieval distribution of description $d$ (Eq.~\ref{eq:retdist}) \\
$m \in \mathbb{N}$ & Observation size (top-$m$ retrieval mass) \\
$\Pop$ & Population of descriptions \\
$\Mem$ & Tool memory (intent strings + evaluated-description history) \\
$\mathcal{P}_{\Mem}$ & KDE posterior over description embeddings in $\Mem$ \\
$\mathcal{S}$ & Answer set (DFSDT) \\
$\tau \in \mathbb{R}^+$ & Temperature \\
$\theta \in [0,1]$ & Early-stop threshold applied to $r(d^*)$ \\
$\alpha \in [0,1]$ & Top-1 similarity weight \\
$\sigma \in \mathbb{R}^+$ & KDE kernel bandwidth \\
$\lambda \in \mathbb{R}^+$ & KL-drift penalty weight \\
$D_{\mathrm{KL}}(\cdot\,\|\,\cdot)$ & KL divergence (posterior drift) \\
$p_{\mathrm{cross}} \in [0,1]$ & Crossover branch probability \\
$N \in \mathbb{N}$ & Population size \\
$G \in \mathbb{N}$ & Number of generations \\
$B \in \mathbb{N}$ & Retrieval budget \\
\bottomrule
\end{tabular}
\end{table}

\section{Memetic Extensions}
\label{app:memetic_extensions}

\subsection{Memetic V2: A Negative Result}
\label{app:memetic_v2}

We also implemented and evaluated a heavier variant, \emph{Memetic V2}, which augments the headline algorithm with an LLM-scored term, temperature annealing, staleness decay, embedding-based crowding, tournament selection, an adaptive crossover rate, and guided mutation. V2 ($N{=}8$, $G{=}3$) underperformed the headline configuration by $13.5$ points on the legacy GPT-4.1-mini grid ($59.9$ vs.\ $73.4$ avg.; Table~\ref{tab:full_passrate}) and was not carried forward to the current-solver experiments. Because several operators change jointly, this negative result does not identify which extension caused the loss.

\subsection{Tool Memory Mechanism}
\label{app:memory}

Tool memory $\Mem$ has two components. The first is a shared dictionary mapping pseudo-tool \emph{intent strings} to generation indices; it persists across all branches of a single DFSDT invocation and, within evolutionary strategies, across generations. The second is the \emph{evaluated-description history}: the embeddings of every description scored so far, appended after each fitness evaluation, which the KL penalty of Eq.~\ref{eq:fitness} reads as a particle approximation of the population's posterior over tool intent. The interaction with the tool database can be understood as \emph{tool exploration}: the database is large and heterogeneous, so previously encountered information guides the search using imperfect signals.

\paragraph{KL posterior-drift penalty.}
The memory penalty (Eq.~\ref{eq:fitness}) places a kernel-density estimate over the normalized embeddings of the descriptions in the history (Gaussian kernel, bandwidth $\sigma$) and scores a candidate by the KL divergence between the particle-weight distributions before and after admitting it, clamped at zero. The closed-form weight update follows the Jensen approximation $D_{\mathrm{KL}} \approx \sum_i w'_i \log (w'_i / w_i)$, plus a fresh-mass term for the new particle. Candidates that crowd already-covered regions of description space redistribute weight and incur drift; novel candidates do not, encouraging search over new areas of the corpus.

\paragraph{Write points.}
\begin{enumerate}[leftmargin=*]
  \item \textbf{After non-evolutionary retrieval} (Single-Pass, IBI, DBD): all pseudo-tool descriptions from the completed call are added to the intent dictionary.
  \item \textbf{After evolutionary finalization}: the ancestor $a$ and the best individual $d^*$ are added with the current generation counter.
  \item \textbf{After each fitness evaluation}: the scored description's embedding is appended to the evaluated-description history, so subsequent candidates are scored against the updated population.
\end{enumerate}

\paragraph{Read points.}
\begin{enumerate}[leftmargin=*]
  \item \textbf{DFSDT duplicate check} (Algorithm~\ref{alg:dfsdt}): suppresses redundant retrieval across sibling branches using SequenceMatcher similarity $\geq 0.82$ over intent strings.
  \item \textbf{Fitness evaluation} (Eq.~\ref{eq:fitness}): the KL term reads the evaluated-description history as the prior posterior and penalizes the drift a candidate induces.
\end{enumerate}

\subsection{Evolutionary Extensions and Framework}
\label{app:evolutionary}

\ourmethod draws on ideas from evolutionary computation, particularly memetic algorithms \citep{guo2025evoprompt}. Table~\ref{tab:evo_mapping} maps \ourmethod concepts to their evolutionary counterparts.

\begin{table}[ht]
\centering
\renewcommand{\arraystretch}{1.15}
\small
\caption{Mapping between \ourmethod concepts and evolutionary computation terminology.}
\label{tab:evo_mapping}
\begin{tabular}{@{}ll@{}}
\toprule
\textbf{\ourmethod Concept} & \textbf{Evolutionary Term} \\
\midrule
Pseudo-tool description & Meme / Gene \\
Set of candidate descriptions & Population \\
LLM refinement & Mutation \\
Retrieval quality score & Fitness \\
Tool database & Environment \\
Population-level voting & Selection \\
Multi-turn feedback (DBD operator) & Local search \\
Scattershot diverse generation & Exploration \\
\bottomrule
\end{tabular}
\renewcommand{\arraystretch}{1.0}
\end{table}

\paragraph{Temperature importance.}
We use temperature $\tau = 1.5$ for population generation to ensure sufficient diversity. Higher temperature promotes solution divergence \citep{li2025divergence}, which is critical for the exploration phase of evolutionary search---\citet{yang2024opro} observe that high temperature ``allows the LLM to more aggressively explore solutions that can be notably different,'' directly motivating our choice. Without sufficient initial diversity, the population converges prematurely and the evolutionary operators (crossover, mutation, selection) have limited material to work with.

\subsection{Strategy Design Space}
\label{app:design_space}

Table~\ref{tab:design_space} compares the key design dimensions across all \ourmethod strategies.
Rather than separate algorithms, the strategies are best read as boundary cases within a single design space defined by these dimensions, with Memetic the most general configuration and the others obtained by disabling operators. In these terms, Scattershot is the boundary case with no fitness-driven selection or local search ($G{=}1$), and Multi-Turn (DBD/IBI) is the boundary case with a single lineage member and no crossover or shared tool-memory across members (effectively population size $N{=}1$), retaining only iterative refinement. We do not treat these as strict degenerate cases but as points in a common space that let us isolate the contribution of each operator.

\begin{table*}[t]
\centering
\renewcommand{\arraystretch}{1.15}
\setlength{\tabcolsep}{4.5pt}
\small
\caption{Strategy design space comparison. LLM cost level is relative per lineage.}
\label{tab:design_space}
\begin{tabular}{@{}lcccccl@{}}
\toprule
\textbf{Strategy} & \textbf{Iterative} & \textbf{Indep.\ lineages} & \textbf{Diverse gen.} & \textbf{Fitness-driven} & \textbf{Pop.\ voting} & \textbf{LLM cost} \\
\midrule
Single-Pass & No & --- & No & No & No & None \\
IBI & Yes & No & No & No & No & Low \\
DBD & Yes & Yes & No & No & No & Low \\
Scattershot & No & Yes & Yes & No & Yes & Medium \\
Memetic & Yes & Yes & Yes & Yes & Yes & High \\
Memetic V2 & Yes & Yes & Yes & Yes & Yes & Highest \\
\bottomrule
\end{tabular}
\renewcommand{\arraystretch}{1.0}
\end{table*}

\section{Earlier-Generation and Extended Empirical Results}
\label{app:legacy_extended}

This section consolidates results that are not part of the headline
claims of the main paper: (i) the complete pass-rate grid for the
\emph{earlier-generation} GPT-4.1-mini solver, retained for historical
comparability; and (ii) the accepted-response wrong-first-tool recovery
diagnostic, whose companion lead appears in the main body. We close
with the evaluation conventions used throughout (\S\ref{app:provenance}).

\subsection{Legacy GPT-4.1-mini Results}
\label{app:legacy_gpt41}

The results in this subsection were obtained with \textbf{GPT-4.1-mini},
an earlier-generation solver that predates the GPT-5.4-mini,
Qwen3.6-35B, and Gemma4-31B configurations reported in the main body.
They are retained here for historical comparability and are
\emph{excluded from all headline claims}.

\paragraph{Legacy StableToolBench pass-rate grid.}
Table~\ref{tab:legacy_gpt41_stb} reports the full StableToolBench
pass-rate grid for the legacy GPT-4.1-mini solver, judged by
GPT-5.4-mini with \texttt{eval\_times}${=}8$ on the 765-task solvable
subset (G1-C~153 / G1-I~163 / G1-T~158 / G2-C~124 / G2-I~106 /
G3-I~61). Row labels are internal run names: \emph{normal} is the
SimCSE Query (static) baseline arm, and \emph{just\_query} the LLM-free
raw-query retrieval arm. The \emph{single\_pass} average is unavailable
because 60 official G3 query IDs were missing from that run.
Note that this table and the legacy grid of
Table~\ref{tab:full_passrate} derive from \emph{different evaluation
passes} (judge protocol and run selection differ), so cells are not
comparable across the two tables; comparisons within each table are
internally consistent, and neither contributes to headline claims.

\begin{table}[t]
\centering
\small
\caption{Legacy \textbf{GPT-4.1-mini} pass rate (\%) on StableToolBench
(judge GPT-5.4-mini, \texttt{eval\_times}${=}8$, 765 solvable). This is
an \emph{earlier-generation} solver, retained for historical
comparability and excluded from headline claims.}
\label{tab:legacy_gpt41_stb}
\begin{tabular}{lccccccc}
\toprule
\textbf{Strategy} & \textbf{G1-C} & \textbf{G1-I} & \textbf{G1-T} & \textbf{G2-C} & \textbf{G2-I} & \textbf{G3-I} & \textbf{Pooled} \\
\midrule
memetic                  & 76.0 & 61.8 & 62.6 & 69.2 & 64.9 & 62.5 & 66.5 \\
scatter\_s5              & 69.0 & 59.5 & 56.4 & 64.5 & 57.8 & 53.5 & 60.8 \\
dbd\_t3                  & 67.1 & 56.0 & 55.0 & 58.0 & 55.8 & 45.4 & 57.5 \\
normal                   & 64.2 & 50.3 & 52.5 & 46.1 & 48.3 & 48.6 & 52.4 \\
just\_query              & 67.6 & 51.7 & 54.7 & 53.8 & 49.8 & 60.3 & 56.3 \\
single\_pass             & 64.3 & 52.3 & 57.2 & 58.9 & 23.6 & 50.5 & \multicolumn{1}{c}{---$^{\dagger}$} \\
\bottomrule
\end{tabular}

\vspace{2pt}
{\footnotesize $^{\dagger}$ Pooled average unavailable: 60 official
G3 query IDs were missing from the legacy \emph{single\_pass} run.}
\end{table}

\paragraph{Legacy ToolRet retrieval grid.}
The per-subset ToolRet retrieval results for the GPT-4.1-mini solver
are already tabulated in Tables~\ref{tab:memetic_code},
\ref{tab:memetic_customized}, and \ref{tab:memetic_web} (Memetic,
per domain).
A strategy-level legacy ToolRet retrieval grid (Query Retrieval /
Single-Pass / Scattershot / DBD / Memetic) analogous to the
main-paper retrieval table is not reported here for this
earlier-generation solver.

\subsection{Current Work and Wall-Clock Measurements}
\label{app:current_latency}

Table~\ref{tab:current_latency} reports the accepted-response
end-to-end measurements for the main GPT-5.4-mini configuration. All
totals include the downstream DFSDT solve.

\begin{table}[t]
\centering
\small
\setlength{\tabcolsep}{4pt}
\caption{Measured GPT-5.4-mini end-to-end token use and wall-clock. Token
totals include the downstream solve; wall-clock is measured
at 40-way benchmark concurrency.}
\label{tab:current_latency}
\begin{tabular}{@{}lccc@{}}
\toprule
\textbf{Strategy} & \textbf{Tokens} & \textbf{Wall-clock} & \textbf{Pass (\%)} \\
\midrule
Single-Pass & 26.1k & 152s ($1.00\times$) & 62.1 \\
Multi-Turn  & 45.2k & 134s ($0.88\times$) & 70.7 \\
Scattershot & 45.6k & 108s ($0.71\times$) & 77.6 \\
Memetic     & 60.5k & 154s ($1.01\times$) & \textbf{84.3} \\
\bottomrule
\end{tabular}
\end{table}

Memetic improves pass rate by $22.2$ points over Single-Pass
($84.3\%$ vs.\ $62.1\%$). Its extra total
tokens are upstream search: the downstream solver core is essentially
constant (Single-Pass vs.\ Memetic, approximately $10.2$k vs.\ $10.7$k
median tokens and $26.1$k vs.\ $26.6$k mean). Population calls run in
parallel, so the extra summed work does not add the same serialized
delay. The wall-clock result is throughput under load; a query run in
isolation would be slower.

\paragraph{Coverage boundaries.}
The four current operating points support incremental comparisons
against Single-Pass, but not a complete cross-strategy, cross-model
latency frontier. Latency is a throughput-under-load measurement at
concurrency $40$, not isolated-query latency. Open-source solver
GPU-hours were not instrumented.

\subsection{Wrong-First-Tool Recovery}
\label{app:recovery_full}

The accepted-response analysis isolates the mechanism after an
incorrect first tool call. First-tool accuracy remains within a
$\sim$5.5-point band across strategies, so \ourmethod's advantage does
not come from choosing a better first tool. It comes from revising the
retrieval frame after execution feedback.

\begin{table}[t]
\centering
\small
\caption{Accepted-response recovery diagnostic on wrong-first-tool G2
tasks: fraction (\%) for which a correct tool appears in the
deduplicated top-5 candidate frame. The dynamic column aggregates
Multi-Turn, Scattershot, and Memetic.}
\label{tab:recovery_accepted}
\begin{tabular}{lrr}
\toprule
\textbf{Split} & \textbf{Frozen root frame} & \textbf{Dynamic re-retrieval} \\
\midrule
G2-category    & $\sim$7  & $\sim$61 \\
G2-instruction & $\sim$23 & $\sim$73 \\
\bottomrule
\end{tabular}
\end{table}

The asymmetry is structural. Static retrieval changes its frame in
\textbf{0 of 2{,}047 invocations}; after a miss, the agent can only
reconsider tools already exposed at the root. Dynamic re-retrieval can
test a revised hypothesis against the full corpus and restore a correct
candidate that was previously absent. This self-correction channel
reduces the pooled unsolved rate from \textbf{$45.9\%$ under static
retrieval to $34.4\%$ under Memetic}. Together with the stable
first-tool accuracy, these results show that Memetic's gains arise from
recovery after feedback rather than from an easier initial choice.

\subsection{Evaluation Conventions}
\label{app:provenance}

For reproducibility we fix two evaluation conventions, applied consistently throughout. \textbf{NDCG mode:} ToolRet retrieval is reported under the \textbf{Global} NDCG protocol (merge qrels and results, then evaluate), matching the per-subset retrieval tables. \textbf{Refinement depth:} description-by-description refinement (DBD / Multi-Turn) uses $T{=}3$ for GPT-5.4-mini and Gemma4-31B; for Qwen3.6-35B the Multi-Turn cell uses $T{=}5$.

\section{Discussion: Outlook}
\label{app:discussion}

We close with one speculative direction, stressing that it is untested and that we make no falsifiable claim on its basis.

\paragraph{Tool chain emergence (outlook).}
We did not analyze chain emergence and make no claim that it occurs; the following is a hypothesis for future work. In multi-step tasks, the agent generates multiple pseudo-tool descriptions that are refined across retrieval episodes. One could conjecture that when these descriptions are refined in parallel, they develop complementary specializations---one description targeting data retrieval while another targets data transformation---and that tool chains could form without explicit orchestration. Whether this happens, and whether it would mirror memeplex formation in cultural evolution, remains untested.

\section{Prompt Templates}
\label{app:prompts}

We provide the key LLM prompt templates used across the \ourmethod pipeline. Template variables are shown as \texttt{\{variable\}}. The special tokens \texttt{\{BEGIN\}} and \texttt{\{END\}} delimit pseudo-tool description blocks that trigger retrieval.

\paragraph{Dynamic Retrieval Protocol.}
The following preamble is prepended to the agent's system prompt in open-domain mode, instructing the agent to use the \texttt{\{BEGIN\}...\{END\}} protocol for on-demand tool retrieval:

\begin{lstlisting}[basicstyle=\ttfamily\scriptsize,breaklines=true,frame=single,xleftmargin=4pt,xrightmargin=4pt]
You should use functions to help handle the real time user queries.
1. Dynamic Function Retrieval: The only function available at the start
   is "Finish". To call any other function, you must first retrieve it
   from a function corpus based on your needs.
2. Retrieval Process: To retrieve the required function, write:
   {BEGIN} [Describe the function you need in detail] {END}
   If you need multiple functions, describe each separately using its
   own {BEGIN}...{END} block. Once written, pause and wait until the
   system retrieves the available functions for you.
3. Task Completion: You must always call "Finish" at the end. The final
   output must be self-contained and complete.
\end{lstlisting}

\paragraph{DBD Refinement.}
Given the current pseudo-tool description, the original ancestor, and exemplar tools from prior retrievals, the LLM refines the description to better match the tool database:

\begin{lstlisting}[basicstyle=\ttfamily\scriptsize,breaklines=true,frame=single,xleftmargin=4pt,xrightmargin=4pt]
You are assisting an autonomous agent that has already decomposed the
task into multiple subtasks. Your role is to refine exactly one function
description for the current subtask only.
Context:
- Global user query: {query}
- Draft function description to refine: {current_desc}
- Retrieved function examples from the database: {lineage_examples}
Goal: Refine the description so it (1) closely matches the style and
structure of the retrieved examples, and (2) is optimized for retrieval.
Output: Exactly one {BEGIN}...{END} block. No commentary.
\end{lstlisting}

\paragraph{Scattershot Diversification.}
Each call generates one diversified variant; the caller invokes this $S$ times in parallel to build a diverse candidate pool:

\begin{lstlisting}[basicstyle=\ttfamily\scriptsize,breaklines=true,frame=single,xleftmargin=4pt,xrightmargin=4pt]
Your role is to generate exactly one diversified function description.
Context:
- Global user query: {query}
- Draft description to diversify (reference only): {ancestor_desc}
- Retrieved function examples: {exemplar_block}
Goal: Generate a new description that (1) preserves the subtask intent,
(2) follows the style of retrieved examples, (3) is optimized for
retrieval, and (4) introduces variation in phrasing or emphasis.
Output: Exactly one {BEGIN}...{END} block. Vary wording each call.
\end{lstlisting}

\paragraph{Memetic Population Seeding.}
Generates the initial population of $N$ diverse pseudo-tool descriptions from a single ancestor:

\begin{lstlisting}[basicstyle=\ttfamily\scriptsize,breaklines=true,frame=single,xleftmargin=4pt,xrightmargin=4pt]
You must propose diversified pseudo-tools for the SAME subtask so a
retriever can search for better-matching tools.
Context:
- Global user query: {query}
- Ancestor pseudo-tool (keep its intent): {ancestor_desc}
- Retrieved exemplar descriptions: {exemplar_block}
Goal: Produce {population_size} alternative pseudo-tool descriptions
that preserve the ancestor intent but vary wording and structure.
Output: Exactly {population_size} {BEGIN}...{END} blocks. No extra text.
\end{lstlisting}

\paragraph{Memetic Crossover.}
Combines two parent descriptions into one child:

\begin{lstlisting}[basicstyle=\ttfamily\scriptsize,breaklines=true,frame=single,xleftmargin=4pt,xrightmargin=4pt]
Create exactly one coherent child description that combines the
strongest constraints, parameters, and intent from both parents.
- Parent A: {parent1_desc}
- Parent B: {parent2_desc}
- Lineage anchor: {ancestor_desc}
Output: Exactly one {BEGIN}...{END} block. No commentary.
\end{lstlisting}

\paragraph{Memetic Mutation.}
Produces one mutated variant while preserving intent:

\begin{lstlisting}[basicstyle=\ttfamily\scriptsize,breaklines=true,frame=single,xleftmargin=4pt,xrightmargin=4pt]
Create exactly one mutated variant of the parent that keeps the same
intent but adjusts wording/constraints to improve retrieval.
- Parent description: {parent_desc}
- Lineage anchor: {ancestor_desc}
Output: Exactly one {BEGIN}...{END} block. No commentary.
\end{lstlisting}